%% file: main.tex
\documentclass[twoside]{article}

\usepackage{hyperref}
\usepackage[export]{adjustbox}
\usepackage[round]{natbib}
\usepackage{
    amsmath,
    amsthm,
    amssymb,
    booktabs,
    bm,
    float,
    geometry,
    pdflscape,
    subcaption,
    tikz,
    tikzscale,
}
\usepackage{aistats2019}
\include{defs}

\usetikzlibrary{bayesnet}

\renewcommand{\bibname}{References}
\renewcommand{\bibsection}{\subsubsection*{\bibname}}
\let\ACMmaketitle=\maketitle
\renewcommand{\maketitle}{\begingroup\let\footnote=\thanks \ACMmaketitle\endgroup}

\begin{document}

\twocolumn[

\aistatstitle{\papertitle}

\aistatsauthor{ Sharad Vikram \And Matthew D. Hoffman \And Matthew J. Johnson }

\aistatsaddress{ U.C. San Diego\thanks{Work done while an intern at Google.} \And Google AI \And Google Brain } ]

\begin{abstract}
In variational autoencoders, the prior on the latent codes $z$ is often treated as an afterthought, but the prior shapes the kind of latent representation that the model learns.
If the goal is to learn a representation that is interpretable and useful, then the prior should reflect the ways in which the high-level factors that describe the data vary.
The ``default'' prior is an isotropic normal, but if the natural factors of variation in the dataset exhibit discrete structure or are not independent, then the isotropic-normal prior will actually encourage learning representations that \emph{mask} this structure.
To alleviate this problem, we propose using a flexible Bayesian nonparametric hierarchical clustering prior based on the time-marginalized coalescent (TMC).
To scale learning to large datasets, we develop a new inducing-point approximation and inference algorithm.
We then apply the method without supervision to several datasets and examine the interpretability and practical performance of the inferred hierarchies and learned latent space.
\end{abstract}

\section{Introduction}

Variational autoencoders \citep[VAEs; ][]{kingma2013autoencoding,rezende2014stochastic} are a popular class of deep latent-variable models. The VAE assumes that observations $x$ are generated by first sampling a latent vector $z$ from some tractable prior $p(z)$, and then sampling $x$ from some tractable distribution $p(x\mid g_\theta(z))$.
For example, $g_\theta(z)$ could be a neural network with weights $\theta$ and $p(x\mid g_\theta(z))$ might be a Gaussian with mean $g_\theta(z)$.

VAEs, like other unsupervised latent-variable models \citep[e.g.; ][]{tipping1999probabilistic,blei2003latent}, can uncover latent structure in datasets.
In particular, one might hope that high-level characteristics of the data are encoded more directly in the geometry of the latent space $z$ than they are in the data space $x$. For example, when modeling faces one might hope that one latent dimension corresponds to pose, another to hair length, another to gender, etc. 

What kind of latent structure will the VAE actually discover?
\citet{hoffman2016elbo} observe that the ELBO encourages the model to make the statistics of the population of encoded $z$ vectors resemble those of the prior, so that $p(z)\approx\mathbb{E}_\mathrm{population}[p(z\mid x)]$.
The prior $p(z)$ therefore plays an important role in shaping the geometry of the latent space.
For example, if we use the ``default'' prior $p(z)=\N(z; 0, I)$, then we are asking the model to explain the data in terms of smoothly varying, completely independent factors \citep{burgess2018understanding}. These constraints may sometimes be reasonable---for example, geometric factors such as pose or lighting angle may be nearly independent and rotationally symmetric. But some natural factors exhibit dependence structure (for example, facial hair length and gender are strongly correlated), and others may have nonsmooth structure (for example, handwritten characters naturally cluster into discrete groups).

In this paper, we propose using a more opinionated prior on the VAE's latent vectors: the time-marginalized coalescent \citep[TMC; ][]{boyles2012time}.
The TMC is a powerful, interpretable Bayesian nonparametric hierarchical clustering model that can encode rich discrete and continuous structure.
Combining the TMC with the VAE combines the strengths of Bayesian nonparametrics (interpretable, discrete structure learning) and deep generative modeling (freedom from restrictive distributional assumptions).

Our contributions are:
\begin{itemize}
    \item We propose a deep Bayesian nonparametric model that can discover hierarchical cluster structure in complex, high-dimensional datasets.
    \item We develop a minibatch-friendly inference procedure for fitting TMCs based on an inducing-point approximation, which scales to arbitrarily large datasets.
    \item We show that our model's learned latent representations consistently outperform those learned by other variational (and classical) autoencoders when evaluated on downstream classification and retrieval tasks.
\end{itemize}

\section{Background}

\subsection{Bayesian priors for hierarchical clustering}
\label{sec:bnhc}

\begin{figure*}[t]
\centering
\begin{subfigure}[t]{0.4\textwidth}
    \centering
    \includegraphics[frame, width=\textwidth]{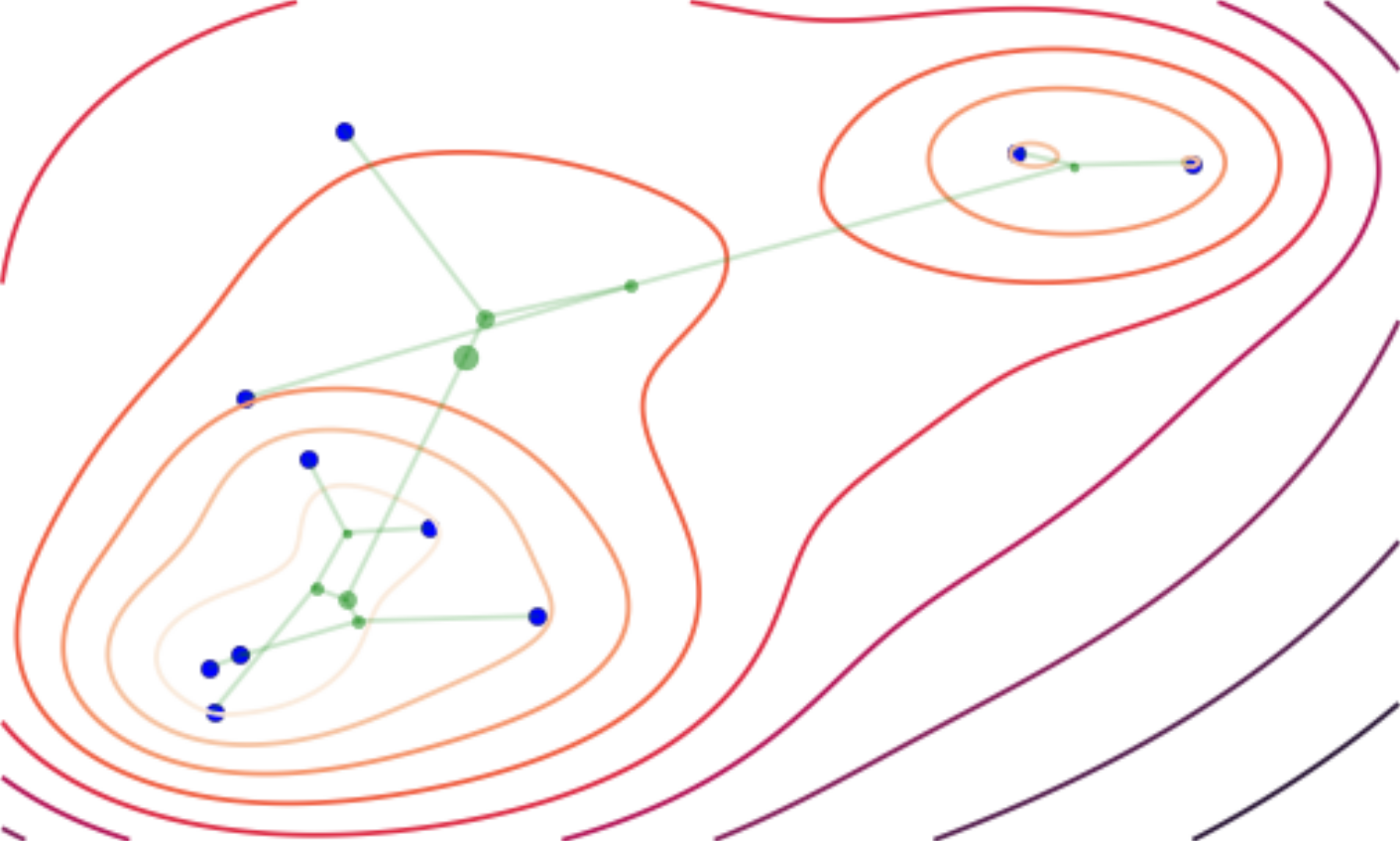}
\end{subfigure}
~
\begin{subfigure}[t]{0.4\textwidth}
    \centering
    \reflectbox{\includegraphics[frame, width=\textwidth]{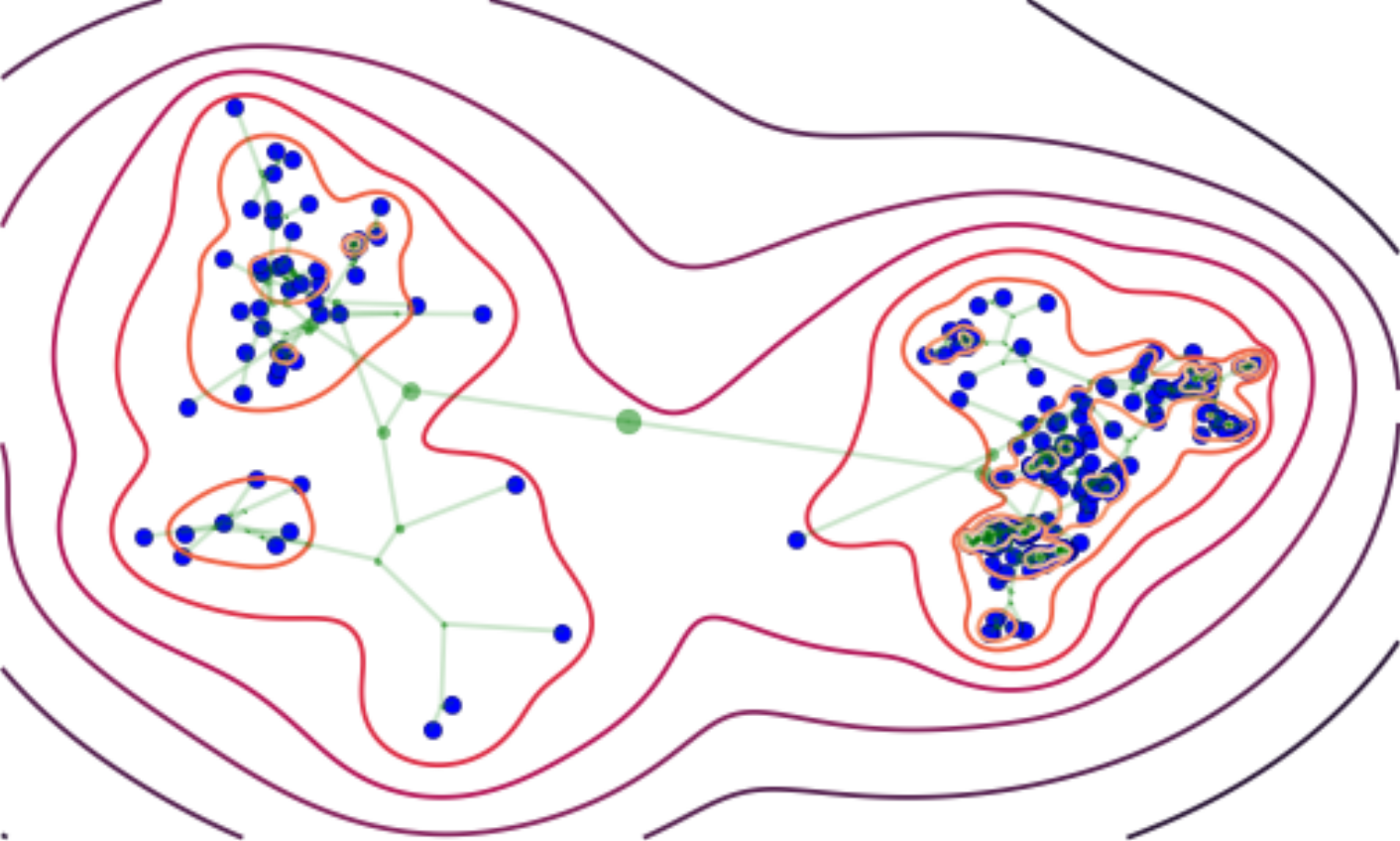}}
\end{subfigure}
\caption{Independent samples from a time-marginalized coalescent (TMC) prior and two-dimensional Gaussian random walk likelihood model (10 and 300 leaves respectively). 
Contours in the plots correspond to posterior predictive density $r(z_{N + 1} \given z_{1:N}, \tau)$.
As the number of leaves grow, the predictive density grows more complex.}
\label{fig:tmc-samples}
\end{figure*}

Hierarchical clustering is a flexible
tool in exploratory data analysis
as trees offer visual, interpretable
summaries of data. Typically,
algorithms for hierarchical clustering are
either agglomerative
(where data are recursively, greedily merged to form
a tree from the bottom-up)
or divisive (where data
are recursively partitioned, forming a tree from the
top-down). Bayesian nonparametric hierarchical clustering (BNHC) 
additionally incorporates uncertainty over tree
structure by introducing
a prior distribution over trees $r(\tau)$ and
a likelihood model for data $r(z_{1:N} \given \tau)$,
with the goal of sampling
the posterior distribution $r(\tau \given z_{1:N})$.\footnote{We use $r$ to denote probability distributions
relating to the TMC and distinguish from $p$ and $q$
distributions used later in the paper.}

In this paper, we focus on
rooted binary trees with $N$ labeled leaves 
adorned with branch lengths,
called \emph{phylogenies}.
Prior distributions over phylogenies
often take the form of a stochastic generative
process in which a tree is built
with random merges, as in the Kingman coalescent \citep{kingman1982coalescent},
or random splits, as in the Dirichlet diffusion tree \citep{neal2003density}.
These nonparametric distributions have
helpful properties, such as exchangeability,
which enable efficient Bayesian inference.
In this paper, we focus on the time-marginalized
coalescent \citep[TMC; ][]{boyles2012time}, which decouples the
distribution over tree structure and branch length,
a property that helps simplify inference down the line.

\subsubsection{Time-marginalized coalescent (TMC)}
The time-marginalized coalescent defines a prior distribution over phylogenies.
A phylogeny $\tau = (V, E, T)$ is a directed rooted full binary tree, with vertex set $V$ and edges $E$, together with time labels $T: V \to [0, \, 1]$ where we denote $t_v = T(v)$.
The vertex set $V$ is partitioned into $N$ leaf vertices $V_\text{leaf}$ and $N-1$ internal vertices $V_\text{int}$, so that $V = V_\text{int} \cup V_\text{leaf}$, and we take $V_\text{leaf} = \{1, 2, \ldots, N\}$ to simplify notation for identifying leaves with $N$ data points.
The directed edges of the tree are encoded in the edge set $E \subset V_\text{int} \times V$, where we denote the root vertex as $v_\text{root}$ and for $v \in V \setminus \{v_\text{root}\}$ we denote the parent of $v$ as $\pi(v) = w$ where $(w, v) \in E$.

The TMC samples a random tree structure $(V, E)$ by a stochastic process in which the $N$ leaves are recursively merged uniformly at random until only one vertex is left.
This process yields the probability mass function on valid $(V, E)$ pairs given by
\begin{equation}
    r(V, E) = \frac{(N - 1)!}{\prod_{v \in V_\text{int}} c(v)} \prod_{i = 1}^{N-1} {i+1 \choose 2}^{-1},
\end{equation}
where $c(v)$ denotes the number of internal vertices in the subtree rooted at $v$.
Given the tree structure, time labels are generated via the stick-breaking process
\begin{equation}
    t_v = \begin{cases} 0 & v = v_\text{root}, \\ 1 & v \in V_\text{leaf}, \\ t_{\pi(v)} - \beta_v (1 - t_{\pi(v)}) & v \in V_\text{int} \setminus \{v_\text{root}\}, \end{cases}
\end{equation}
where $\beta_v \iid \sim \mathrm{Beta}(a, b)$ for $v \in V$. These time labels encode a branch length $t_v - t_{\pi(v)}$ for each edge $e = (\pi(v), v) \in E$. We denote the overall density on phylogenies with $N$ leaves as $\mathrm{TMC}_N(\tau; a, b)$.

Finally, to connect the TMC prior to data in $\mathbb{R}^d$, we define a likelihood model $r(z_{1:N} \given \tau)$ on $N$ data points, with $z_n$ corresponding to the leaf vertex $n \in V_\text{leaf}$.
We use a Gaussian random walk (GRW), where for each vertex $v \in V$ a location $z_v \given z_{\pi(v)}$ is sampled according to a Gaussian distribution centered at its parent's location with variance equal to the branch length,
\begin{equation}
    z_v \given z_{\pi(v)} \sim \N(z_{\pi(v)}, (t_v - t_{\pi(v)})I), \quad v \in V \setminus \{v_\text{root}\},
    \notag
\end{equation}
and we take $z_{v_\text{root}} \sim \N(0, I)$.
As a result of this choice, we can exploit the Gaussian graphical model structure to efficiently marginalize out the internal locations $z_v$ associated with internal vertices $v \in V_\text{int}$ and evaluate the resulting marginal density $r(z_{1:N} \given \tau)$. For details
about this marginalization, please refer to \autoref{sec:algorithm-details}.
The final overall density is written as
\begin{equation}
    r(z_{1:N}, \tau) = \mathrm{TMC}_N(\tau; a, b)r(z_{1:N} \given \tau).
\end{equation}
For further details and derivations related to the TMC,
please refer to \citet{boyles2012time}.

\subsubsection{TMC posterior predictive density}
The TMC with $N$ leaves and a GRW likelihood model can be a prior on a set of $N$ hierarchically-structured data, i.e. data that correspond to nodes with small tree distance should have similar location values.
In addition, it also acts as a density from which we can sample new data.
The posterior predictive density $r(z_{N + 1} \given z_{1:N}, \tau)$ is easy to sample thanks to the exchangeability of the TMC.

To sample a new data point $z_{N + 1}$, we select a branch (edge) and a time to attach a new leaf node.
The probability $r(e_{N+1} \given V, E)$ of selecting branch $e_{N+1}$ is proportional to the probability under the TMC prior of the tree with a new leaf attached to branch $e_{N+1}$.
The density $r(t_{N+1} \given e_{N+1}, V, E)$ for a time label $t_{N+1}$ is determined by the stick-breaking process (see \autoref{sec:algorithm-details} for details).
Both of these probabilities are easy to calculate and sample due to the exchangeability of the TMC.

The new location $z_{N + 1}$ can be sampled from  $r(z_{N + 1} \given e_{N + 1}, t_{N + 1}, \tau)$, which is
the Gaussian distribution that comes out of the GRW likelihood model.
Pictured in \autoref{fig:tmc-samples} are samples from a TMC prior and GRW likelihood, where contours correspond to $r(z_{N + 1} \given z_{1:N}, \tau)$.
In addition to modeling hierarchical structure, the TMC is a flexible nonparametric density estimator.

\subsubsection{TMC inference}

The posterior distribution $r(\tau \given z_{1:N})$
is analytically intractable due to the normalization constant
$r(z_{1:N})$ involving a sum over all tree structures,
but it can be approximately sampled via Markov chain Monte-Carlo (MCMC) methods.
We utilize the Metropolis-Hastings
algorithm with a subtree-prune-and-regraft (SPR)
proposal distribution \citep{neal2003density}. An SPR proposal
picks a subtree uniformly at random from $\tau$
and detaches it.
It is then attached back on the tree
to a branch and time picked uniformly at random.
The Metropolis-Hastings acceptance probability is efficient to compute because the joint density $r(\tau, z_{1:N})$ can be evaluated using belief propagation to marginalize the latent values at internal nodes of $\tau$, and many of the messages can be cached.
See \autoref{sec:algorithm-details} for details.

\subsection{Variational autoencoder}
The variational autoencoder (VAE) is a generative model
for a dataset $x_{1:N}$
wherein latent vectors $z_{1:N}$ are sampled
from a prior distribution
and then individually passed into
a neural network observation model with parameters $\theta$,
\begin{equation}
    \begin{split}
        z_{1:N} \sim p(z_{1:N}),
        \qquad
        x_n \given z_n \sim p_\theta(x_n \given z_n),
    \end{split}
\end{equation}
We are interested in the posterior distribution
$p(z_n \given x_n)$, which is not analytically tractable
but can be approximated with a variational distribution
$q_\phi(z_n \given x_n)$, typically a neural network
that outputs parameters of a Gaussian distribution.
The weights of the approximate posterior can be learned
by optimizing the evidence-lower bound (ELBO),
\begin{equation}
    \begin{split}
    \L[q] &\triangleq \E_q \left[\log \frac{p_\theta(x_{1:N}, z_{1:N})}{\prod_n q_\phi(z_n \given x_n)}\right] \\
    \end{split}
\end{equation}
The parameters of the model, $\theta$ and $\phi$,
are learned via stochastic gradient ascent on
the ELBO, using the reparametrization trick
for lower variance gradients \citep{kingma2013autoencoding, rezende2014stochastic}.

\section{The TMC-VAE}
\label{sec:tmc-vae}

The choice of prior distribution in the VAE significantly affects the autoencoder and resulting latent space.
The default standard normal prior, which takes $z_n \iid\sim \N(0, I)$, acts as a regularizer on an otherwise unconstrained autoencoder, but can be restrictive and result in overpruning \citep{burda2015importance}.
Extremely flexible, learnable distributions like masked autoregressive flow (MAF) priors \citep{papamakarios2017masked} enable very rich latent spaces, but don't encode any interpretable bias for organizing the latent space (except perhaps smoothness).

In this paper, we explore the TMC prior for the VAE, which could potentially strike a sweet spot between restrictive and flexible priors.
We generate the latent values $z_{1:N}$ of a VAE according to the TMC prior, then generate observations $x_{1:N}$ using a neural network observation model,
\begin{gather}
        \tau \sim \mathrm{TMC}_N(\tau; a, b), \\
        z_{1:N} \given \tau \sim r(z_{1:N} \given \tau),
        \qquad
        x_n \given z_n \sim p_\theta(x_n \given z_n).
\end{gather}
The TMC-VAE is a coherent
generative process that captures
discrete, interpretable structure in the latent space.
A phylogeny not only has an intuitive inductive bias,
but can be useful for
exploratory data analysis
and introspecting the latent space itself.

Consider doing inference in this model:
first assume variational distributions 
$q_\phi(z_n \given x_n)$ (as in the VAE)
and $q(\tau)$, which results in the
ELBO,
\begin{equation}
    \begin{split}
    \L[q] &= \E_q\left[\log \frac{p(\tau, z_{1:N}, x_{1:N})}{q(\tau)\prod_n q(z_n \given x_n)}\right] \\
    \end{split}
\end{equation}
For fixed $q_\phi(z_n \given x_n)$, we can
sample the optimal $q^*(\tau)$,
\begin{equation}
    q^*(\tau) \propto \exp\{\E_q\left[\log p(\tau, z_{1:N}, x_{1:N})\right]\}
\end{equation}
Because $p(z_{1:N} \given \tau)$ is jointly Gaussian (factorizing
according to tree structure) and $q_\phi(z_n \given x_n)$ is Gaussian, 
expectations with respect to $z_{1:N}$
can move into $\log p(\tau, z_{1:N}, x_{1:N})$. This
enables sampling the expected joint likelihood $\E_q\left[\log p(\tau, z_{1:N})\right]$
using SPR Metropolis-Hastings.
However, optimizing this ELBO is problematic.
$p(z_{1:N} | \tau)$
does not factorize independently, so computing
unbiased gradient estimates from minibatches is impossible
and requires evaluating all the data.
Furthermore, the TMC is limiting
from a computational perspective.
Since a phylogeny has as many leaves
as points in the dataset, 
belief propagation over internal nodes
of the tree slows down linearly as 
the size of the dataset grows.
In addition,
SPR proposals mix very slowly for large trees.
We found these limitations 
make the model impractical for datasets
of more than 1000 examples.

In the next section, we address these computational
issues, while retaining
the interesting properties of the TMC-VAE.

\begin{figure*}[t]
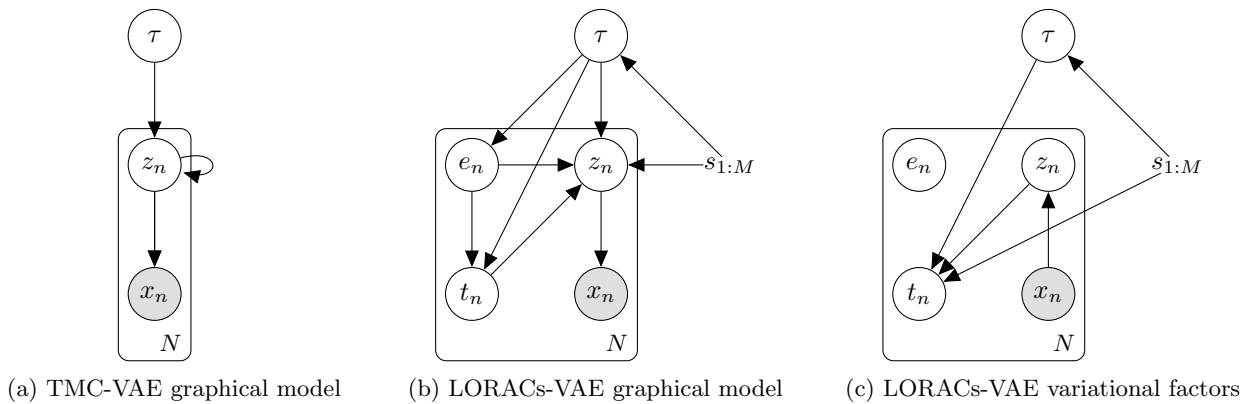

\centering
\begin{subfigure}[t]{0.3\textwidth}
    \centering
    \includegraphics{tikz/ltmc.tikz}
    \caption{TMC-VAE graphical model}
\end{subfigure}
\begin{subfigure}[t]{0.34\textwidth}
    \centering
    \includegraphics{tikz/iltmc.tikz}
    \caption{\acronym-VAE graphical model}
\end{subfigure}
\begin{subfigure}[t]{0.34\textwidth}
    \centering
    \includegraphics{tikz/iltmc-v.tikz}
    \caption{\acronym-VAE variational factors}
\end{subfigure}
\caption{Graphical models and variational
approximations for TMC models described in the paper}
\label{fig:graphical-models}
\end{figure*}

\section{\acronym\;prior for VAEs}

In this section, we introduce
a novel approximation to the TMC
prior, which preserves many 
desirable properties
like structure and interpretability,
while being computationally viable.
Our key idea is to use a set of
learned \emph{inducing points}
as the leaves of the tree in the latent
space, analogous to inducing-input
approximations for Gaussian processes \citep{snelson2006sparse}. 
In this model, latent vectors $z_{1:N}$
are not directly hierarchically clustered,
but are rather independent samples
from the induced posterior predictive density
of a TMC.
We call this the \acronymexplanation\;(\emph{\acronym}, pronounced ``\acronympronunciation'') prior.

To define the \acronym\;prior $p(\tau, z_{1:N})$, we first define an auxiliary TMC distribution $r(\tau, s_{1:M})$ with $M$ leaf locations $s_{1:M}$.
We treat $s_{1:M}$ as a set of learnable free parameters, and define the conditional $r(\tau \given s_{1:M})$ as the \acronym\;prior on phylogenies $\tau$:
\begin{equation}
  p(\tau ; s_{1:M}) \triangleq r(\tau \given s_{1:M}).
\end{equation}
That is, we choose the prior on phylogenies $\tau$ to be the posterior distribution of a TMC with pseudo-observations $s_{1:M}$.
Next, we define the \acronym\;prior on locations $z_n\mid \tau$ as a conditionally independent draw from the predictive distribution $r(s_{M+1} \given \tau, s_{1:M})$, writing the sampled attachment branch and time as $e_n$ and $t_n$, respectively:
\begin{gather}
    p(e_n, t_n \given \tau)
    \triangleq r(e_{M+1} = e_n, t_{M+1} = t_n \given \tau),
    \notag
    \\
    p(z_n \given e_n, t_n, \tau)
    \triangleq r(s_{M+1} = z_n \given e_n, t_n, \tau, s_{1:M}).
\end{gather}
To complete the model, we use an observation likelihood parameterized by a neural network, writing
\begin{equation}
    x_n \given z_n \sim p_\theta(x_n \given z_n).
\end{equation}
By using the learned inducing points $s_{1:M}$, we avoid the main difficulty of inference in the TMC-VAE of Section~\ref{sec:tmc-vae}, namely the need to do inference over all $N$ points in the dataset.
Instead, dependence between datapoints is mediated by the set of inducing points $s_{1:M}$, which has a size independent of $N$.
As a result, with the \acronym\;prior, minibatch-based learning becomes tractable even for very large datasets.
The quality of the approximation to the TMC-VAE can be tuned by adjusting the size of $M$.

However, this technique presents its own inference challenges.
Sampling the optimal variational factor $q^*(\tau)$
is no longer an option as it was in the TMC-VAE:
\begin{equation}
\begin{split}
    q^*(\tau; s_{1:M}) 
    &\textstyle\propto \exp\{\E_q\left[\log p(\tau, z_{1:N}, x_{1:N})\right]\} \\
    &\textstyle\propto \exp\{\log p(\tau ; s_{1:M}) + \sum_n \E_q\left[p(z_n \given e_n, t_n, \tau)\right]\} \\
    &\textstyle\propto \exp\{\log \mathrm{TMC}_N(\tau; a, b)
    \\ &\textstyle\qquad\quad + \sum_{m = 1}^M \log r(s_m \given s_{1:m - 1}, \tau)
    \\ &\textstyle\qquad\quad + \sum_n \E_q\left[\log p(z_n \given e_n, t_n, \tau)\right]\}.
\end{split}
\end{equation}
This term has a sum over $N$ expectations; therefore computing this likelihood
for the purpose of MCMC
would involve passing the entire dataset through a neural network.
Furthermore, the normalizer for this likelihood is intractable,
but necessary for computing gradients w.r.t $s_{1:M}$.
We therefore avoid using the optimal $q^*(\tau; s_{1:M})$
and set
$q(\tau; s_{1:M})$ to the prior.
This has the additional computational advantage of cancelling out
the $\mathbb{E}_q[\log p(\tau)]$ term in the ELBO, which also has an intractable normalizing constant.
If the inducing points are chosen so that they contain most of the information about the hierarchical organization of the dataset, then the approximation $p(\tau \given z)\approx r(\tau \given s_{1:M})=p(\tau)$ will be reasonable.

We also fit the variational factors
$q(e_n)$,
$q_{\xi}(t_n \given e_n, z_n; s_{1:M})$,
and
$q_\phi(z_n \given x_n)$.
The factor for attachment times,
$q_{\xi}(t_n \given e_n, z_n; s_{1:M})$, is a
recognition network that outputs
a posterior over attachment times for a particular branch.
Since the $q(\tau; s_{1:M})$ and $p(\tau; s_{1:M})$ terms
cancel out, we obtain the following ELBO (some notation suppressed for simplicity):
\begin{equation}
    \begin{split}
    \L[q] &\triangleq \E_q \left[\log \frac{\prod_n p(e_n, t_n \given \tau) p(z_n \given e_n, t_n, \tau)p(x_n \given z_n)}{\prod_n q(e_n)q(t_n \given e_n, z_n)q(z_n \given x_n)}\right].
    \end{split}
\end{equation}
This ELBO can be optimized by first computing
\begin{equation}
    q^*(e_n) = \exp\left\{\E_q\left[\log p(e_n \given t_n, z_n, \tau; s_{1:M})\right]\right\}
\end{equation}
and computing gradients with respect to $\theta$, $s_{1:M}$,
$\phi$, and $\xi$ using a Monte-Carlo estimate of the ELBO
using samples from $q(\tau; s_{1:M})$, $q^*(e_n)$, $q_\phi(z_n \given x_n)$, and $q_{\xi}(t_n \given e_n, z_n; s_{1:M})$.
The factor $q(\tau; s_{1:M})$ can
be sampled using vanilla SPR Metropolis-Hastings. 
The detailed inference procedure can be found in \autoref{sec:algorithm-details}.

\section{Related work}
As mentioned above, \acronym\;connects various ideas in the literature, including Bayesian nonparametrics \citep{boyles2012time}, inducing-point approximations \citep[e.g.; ][]{snelson2006sparse,tomczak2018vae},
and amortized inference \citep{kingma2013autoencoding,rezende2014stochastic}.

Also relevant is a recent thread of efforts to endow VAEs with the interpretability of graphical models \citep[e.g.; ][]{johnson2016svae,lin2018variational}.
In this vein, \citet{goyal2017nonparametric} propose using a different Bayesian nonparametric tree prior, the nested Chinese restaurant process (CRP) \citep{griffiths2004hierarchical}, in a VAE.
We chose to base \acronym\;on the TMC instead, as the posterior predictive distribution of an nCRP is a finite mixture, whereas the TMC's posterior predictive distribution has more complex continuous structure.
Another distinction is that \citet{goyal2017nonparametric} only consider learning from pretrained image features, whereas our approach is completely unsupervised.

\section{Results}
In this section, we 
analyze properties of the
\acronym\;prior,
focusing on
qualitative aspects, like exploratory data analysis and interpretability,
and quantitative aspects, like few-shot classification and information retrieval.
\paragraph{Experimental setup} We evaluated
the \acronym\;prior on three separate
datasets: dynamically binarized MNIST \citep{mnist}, Omniglot \citep{omniglot},
and CelebA \citep{celeba}. 
For all three experiments,
we utilized convolutional/deconvolutional encoders/decoders
and a 40-dimensional
latent space (detailed architectures can be found
in \autoref{sec:implementation-details}).
We used 200, 1000, and 500 inducing points for MNIST, Omniglot, and CelebA respectively 
with TMC parameters $a = b = 2$.
$q_{\xi}(t_n \given e_n, z_n; s_{1:M})$ was a
two-layer 500-wide neural network
with ReLU activations that output parameters of a
logistic-normal distribution over stick size
and all parameters were optimized with Adam \citep{kingma2015adam}.
Other implementation details can be found in \autoref{sec:implementation-details}.

\subsection{Qualitative results}

A hierarchical clustering in the latent space
offers a unique opportunity for interpretability
and exploratory data analysis,
especially when the data are images.
Here are some methods for users to obtain
useful data summaries and explore a dataset.

\begin{figure}[t]
\centering
\includegraphics[width=0.5\textwidth]{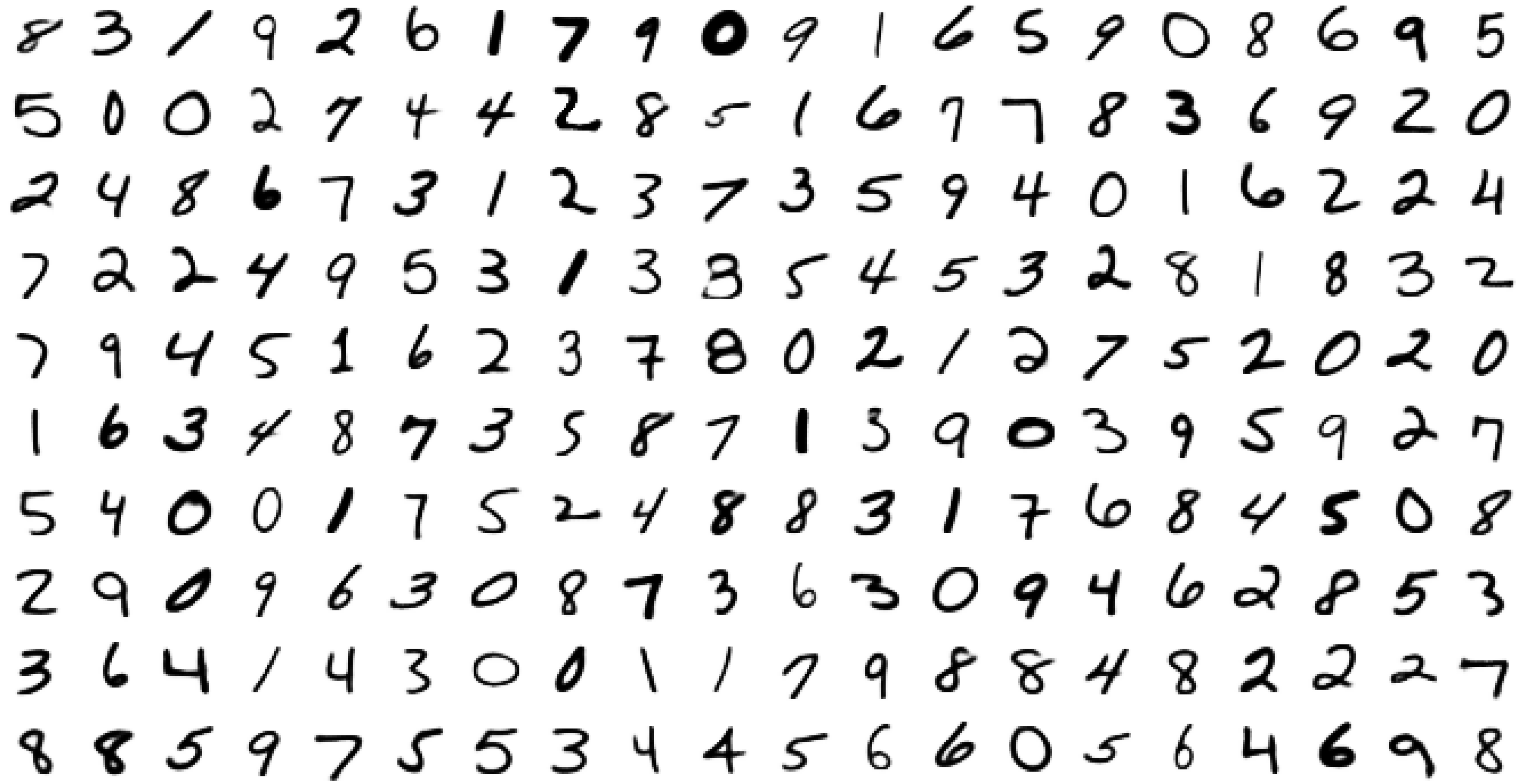}
\caption{Learned inducing points for a \acronym(200) prior on MNIST.}
\label{fig:mnist-inducing}
\end{figure}

\paragraph{Visualizing inducing points}
We first inspect the learned inducing points $s_{1:M}$
by passing them through the decoder.
Visualized in \autoref{fig:mnist-inducing}
are the 200 learned inducing points for MNIST.
The inducing points are all unique
and are cleaner than pseudo-input reconstructions from VampPrior (shown in
\autoref{fig:mnist-vamp-inducing-outputs}).
Inducing points can help summarize a
dataset, as visualizations of the latent space
indicate they spread out and cover the data
(see \autoref{fig:mnist-tsne-tree}). Inducing points
are also visually unique and sensible
in Omniglot and CelebA (see
\autoref{fig:omniglot-inducing-points} and \ref{fig:celeba-inducing-points}).

\begin{figure}[t]
\centering
\begin{subfigure}[t]{0.2\textwidth}
\centering
\includegraphics[width=\textwidth]{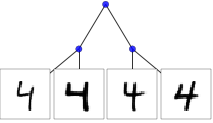}
\caption{MNIST}
\end{subfigure}
\begin{subfigure}[t]{0.2\textwidth}
\centering
\includegraphics[width=\textwidth]{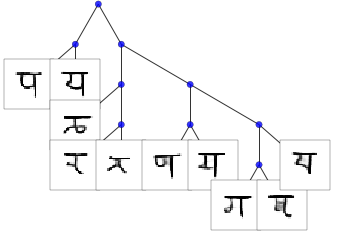}
\caption{Omniglot}
\end{subfigure}
\begin{subfigure}[t]{0.2\textwidth}
\centering
\includegraphics[width=\textwidth]{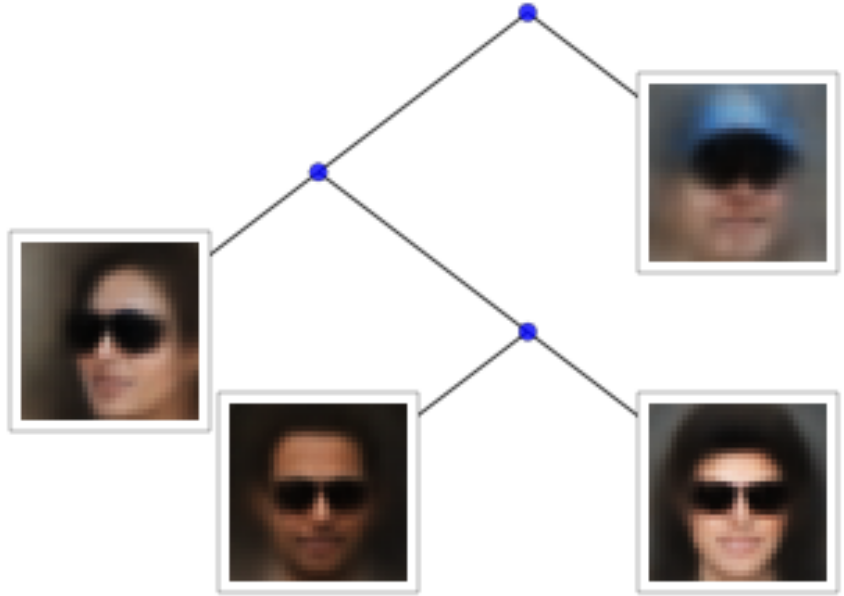}
\caption{CelebA}
\end{subfigure}
\caption{An example learned subtree from a sample of $q(\tau; s_{1:M})$ for each dataset. 
Leaves are visualized by passing inducing points throught the decoder.}
\label{fig:subtrees}
\end{figure}

\paragraph{Hierarchical clustering}
We can sample
$q(\tau ; s_{1:M})$ to obtain
phylogenies
over the inducing points,
and can visualize these clusterings
using the decoded inducing points;
subtrees from a sample in each dataset
are visualized in \autoref{fig:subtrees}.
In MNIST, we find large subtrees
correspond to the discrete classes
in the dataset. In Omniglot,
subtrees sometimes correspond to language groups
and letter shapes. In CelebA,
we find subtrees sometimes correspond
to pose or hair color and style.

\begin{figure}[t]
\centering
\includegraphics[width=0.5\textwidth]{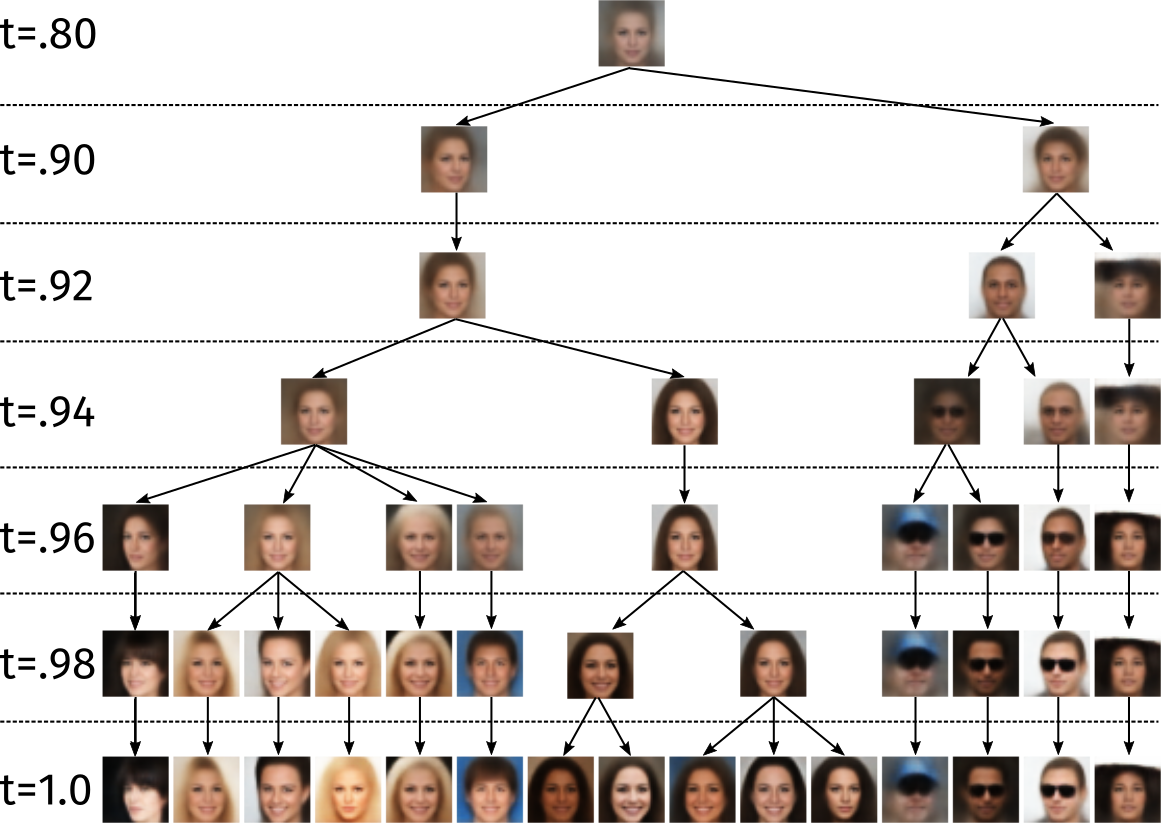}
\caption{The evolution of a CelebA over a subset of inducing points. We create
this visualization by taking slices of the tree at particular times
and looking at the latent distribution 
at each of the sliced branches.}
\label{fig:celeba-evolution}
\end{figure}

We can further use the time at each internal node
to summarize the data at many levels of granularity.
Consider ``slicing'' the hierarchy
at a particular time $t$ by
taking every branch $(\pi(v), v) \in E$ with $t_{\pi(v)} \leq t < t_v$
and computing the corresponding expected Gaussian random walk value at time $t$.
At times closer to zero, we slice
fewer branches and are closer to the root
of the hierarchy, so the value at the slice
looks more like the mean of the data.
In \autoref{fig:celeba-evolution}, 
we visualize this process over a subset
of the inducing points of CelebA.
Visualizing the dataset in this way
reveals cluster structure at 
different granularities
and offers an evolutionary interpretation
to the data, as leaves that coalesce 
more ``recently'' are likely to be closer in 
the latent space.

Although the hierarchical clustering is only
over inducing points, we can still visualize
where real data belong on the hierarchy
by computing $q^*(e_n)$ and
attaching the data to the tree.
By doing this for many points of data, and removing
the inducing points from the tree,
we obtain an induced hierarchical clustering.

\begin{figure}[h]
\centering
\begin{subfigure}[t]{0.45\textwidth}
\centering
\includegraphics[width=\textwidth]{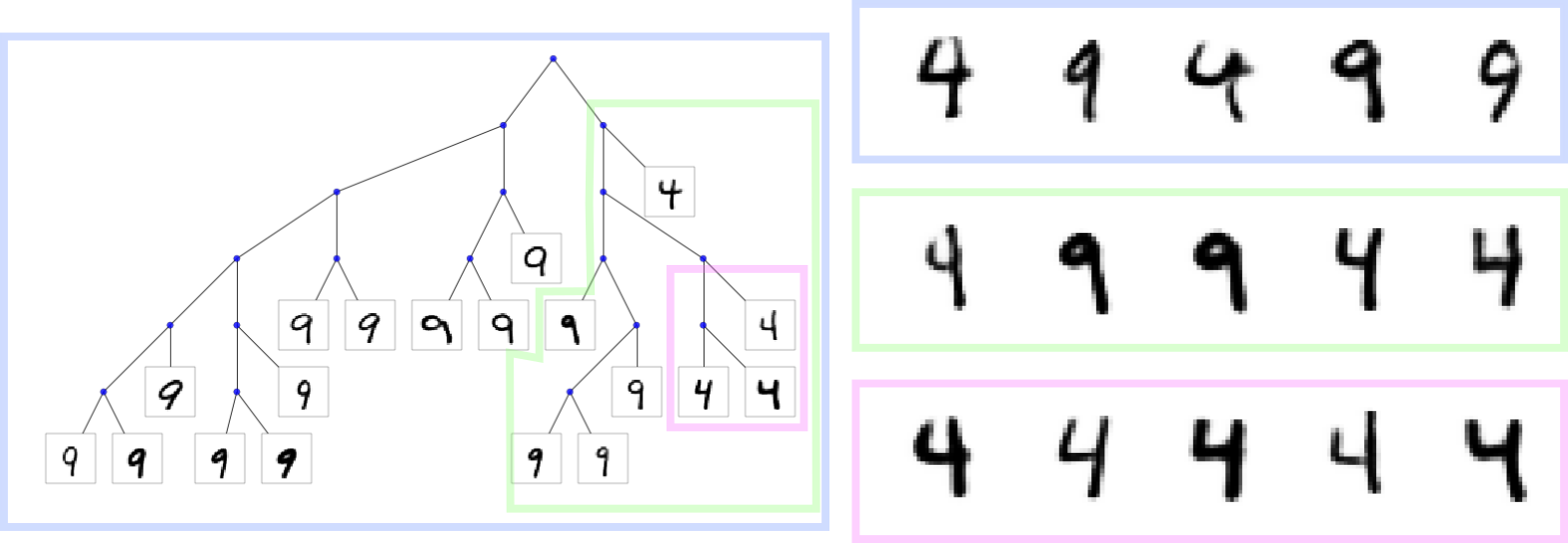}
\caption{MNIST}
\end{subfigure}
\begin{subfigure}[t]{0.45\textwidth}
\centering
\includegraphics[width=\textwidth]{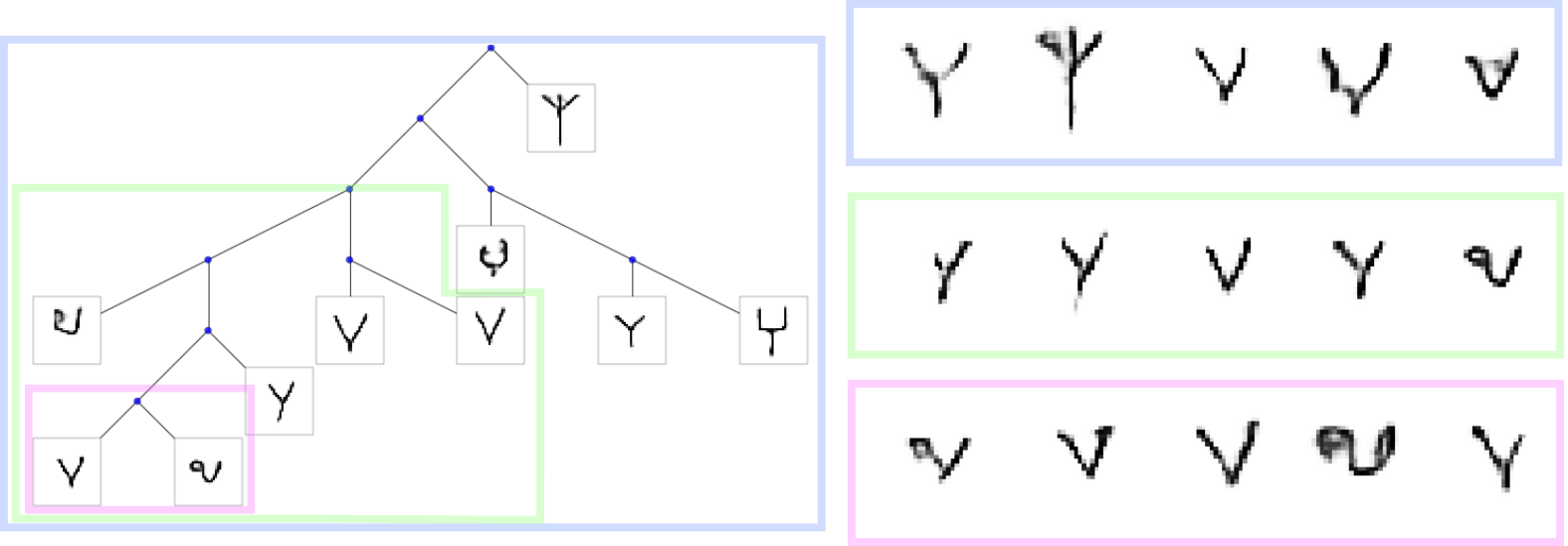}
\caption{Omniglot}
\end{subfigure}
\caption{Conditional samples from subtrees.}
\label{fig:subtree-samples}
\end{figure}

\begin{figure}[h]
\centering
\begin{subfigure}[t]{0.45\textwidth}
\centering
\includegraphics[frame, width=\textwidth]{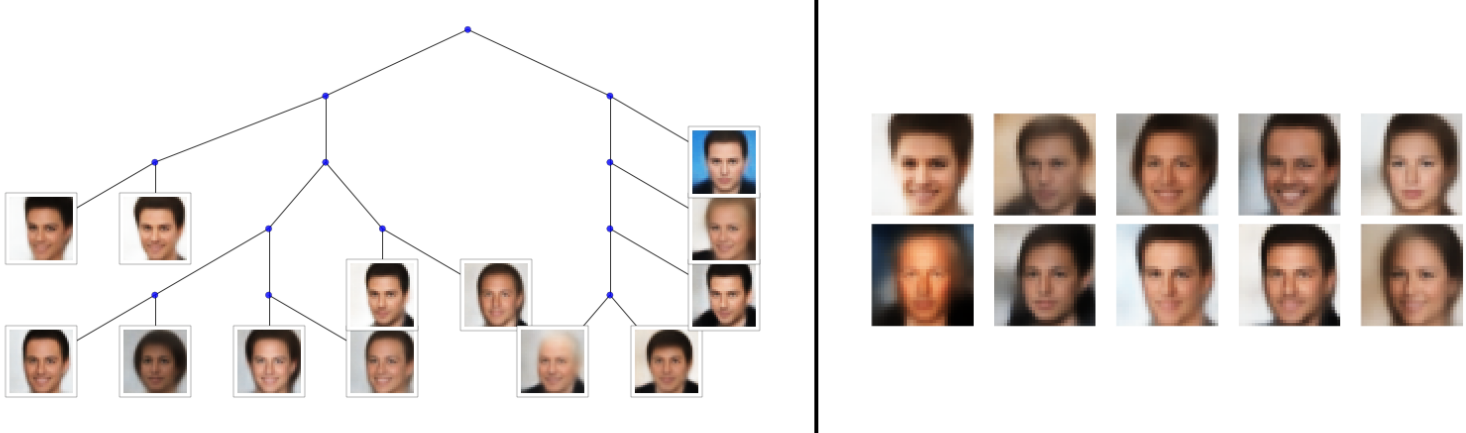}
\end{subfigure}
\begin{subfigure}[t]{0.45\textwidth}
\centering
\includegraphics[frame, width=\textwidth]{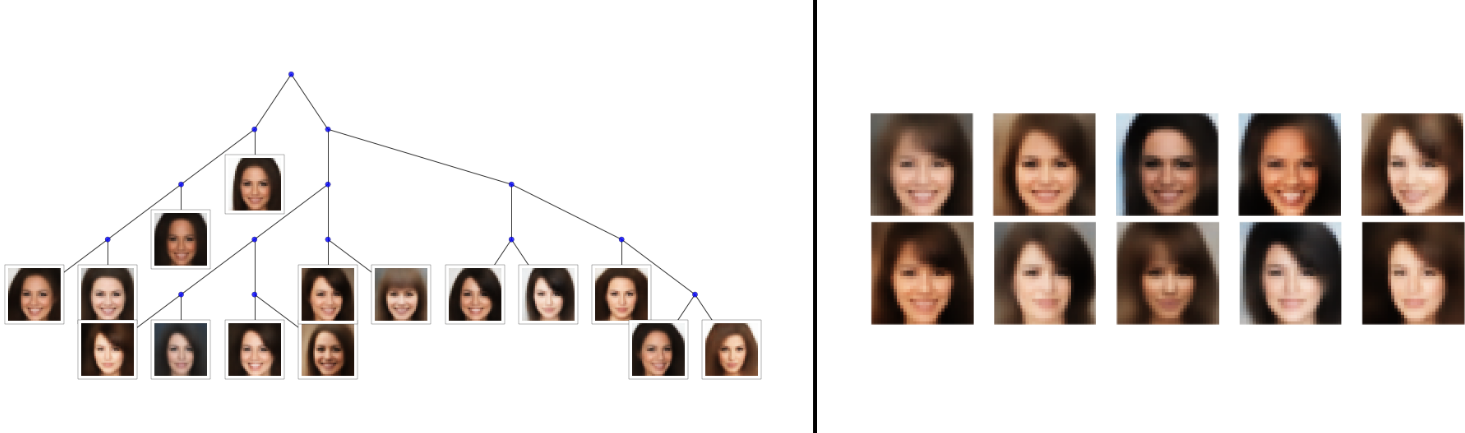}
\end{subfigure}
\caption{Samples from subtrees of CelebA.}
\label{fig:celeba-samples}
\end{figure}

\paragraph{Generating samples} 
Having fit a generative model to our data,
we can visualize samples from the model.
Although we do not expect the samples
to have fidelity and sharpness comparable
to those from GANs or state-of-the-art
decoding networks \citep{radford2015unsupervised, salimans2017pixelcnn++},
sampling with the
\acronym\;prior can help us understand the latent space.
To draw a sample from a TMC's posterior predictive density,
we first sample a branch and time, assigning
the sample a place in the tree.
This provides each generated sample a \emph{context},
i.e., the branch and subtree it was generated from.
However, learning a \acronym\;prior allows us to
conditionally sample in a novel way. By restricting
samples to a subtree, we can generate samples
from the support of the posterior predictive density
limited to that subtree. This enables
conditional sampling at many levels of
the hierarchy. We visualize examples
of this in \autoref{fig:subtree-samples}
and \autoref{fig:celeba-samples}.

\subsection{Quantitative results}

We ran experiments designed to
evaluate the usefulness of the \acronym's
learned latent space for downstream tasks. We 
compare the \acronym\;prior
against a set of baseline priors
on three different tasks:
few-shot classification,
information retrieval,
and generative modeling.
Our datasets are
dynamically binarized MNIST and Omniglot 
(split by instance) and
our baselines are representations
learned with the same encoder-decoder
architecture and latent dimensionality\footnote{Following the defaults
in the author's reference implementation, we evaluated
DVAE\# on statically binarized MNIST with smaller neural networks, but
with a higher-dimensional latent space.} but substituting
the following prior distributions over $z$:
\begin{itemize}
    \setlength\itemsep{0.2em}
    \item No prior
    \item Standard normal prior
    \item VampPrior \citep{tomczak2018vae} - 500 pseudo-inputs for MNIST, 1000 for Omniglot
    \item DVAE$\sharp$ \citep{vahdat2018dvae} - latent vectors are 400-dimensional, formed from concatenating binary latents, encoder and decoder are two-layer feed-forward networks with ReLU nonlinearities
    \item Masked autoregressive flow \citep[MAF; ][]{papamakarios2017masked} - two layer, 512 wide MADE
\end{itemize}
\begin{figure}[t]
\centering
\begin{subfigure}[t]{0.35\textwidth}
    \centering
    \includegraphics[width=\textwidth]{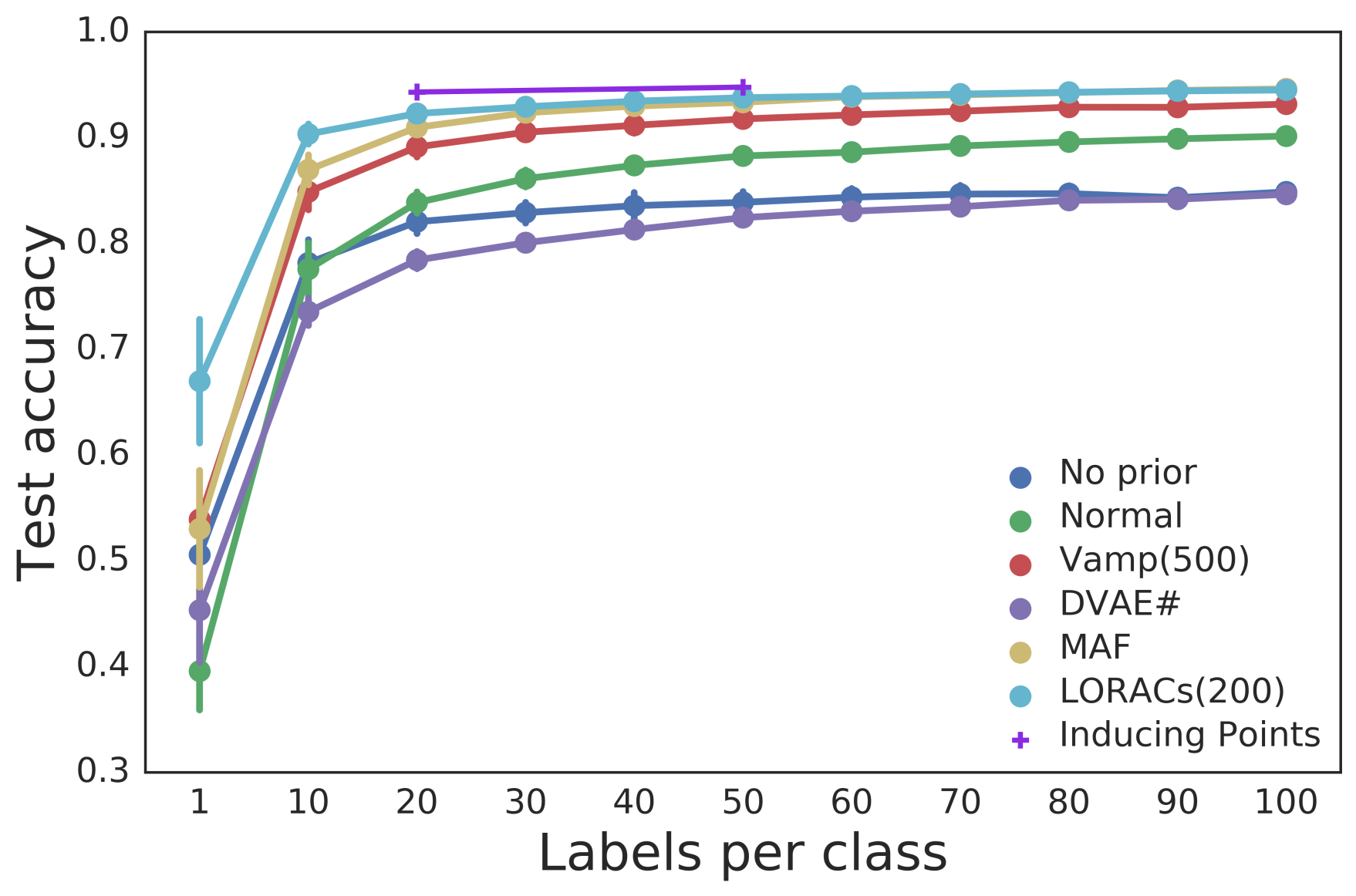}
    \caption{MNIST}
\end{subfigure}
\begin{subfigure}[t]{0.33\textwidth}
    \centering
    \includegraphics[width=\textwidth]{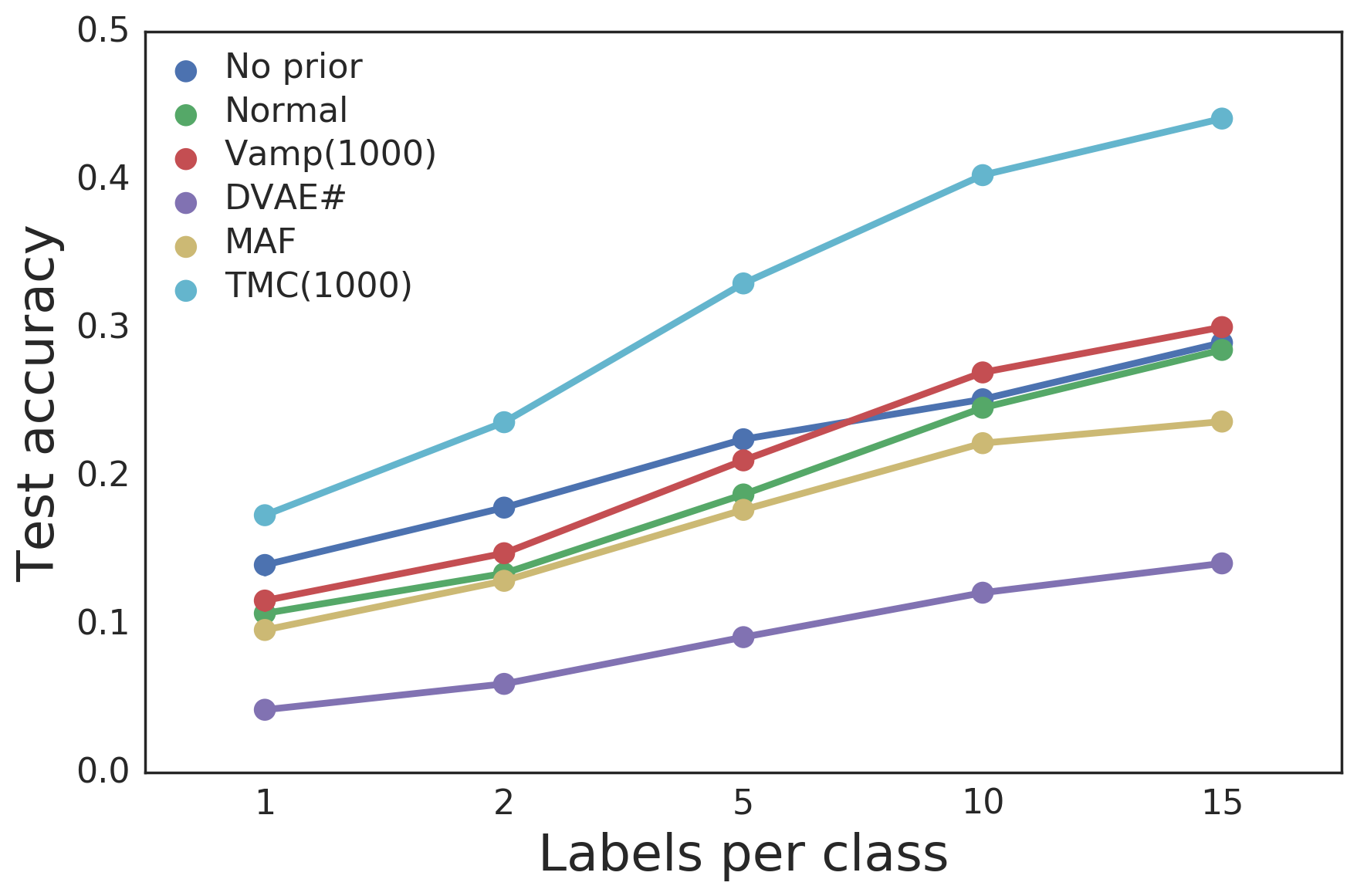}
    \caption{Omniglot}
\end{subfigure}
\caption{Few-shot classification results}
\vspace{-0.5cm}
\label{fig:fewshot}
\end{figure}
\paragraph{Few-shot classification}
In this task,
we train a
classifier with
varying numbers of labels and measure
test accuracy. We pick
equal numbers of labels per class
to avoid imbalance and we use
a logistic regression classifier
trained to convergence to avoid
adding unnecessary degrees of freedom to the experiment.
We replicated the experiment across
20 randomly chosen label sets for MNIST
and 5 for Omniglot.
The test accuracy on
these datasets is visualized in 
\autoref{fig:fewshot}. For MNIST, we 
also manually labeled inducing points and
found that training a classifier on 200 and 500 inducing points
achieved significantly better test accuracy than 
randomly chosen labeled points, hinting that
the \acronym\;prior has utility in an active learning setting.

\begin{figure}[H]
\centering
\begin{subfigure}[t]{0.2\textwidth}
    \centering
    \includegraphics[width=\textwidth]{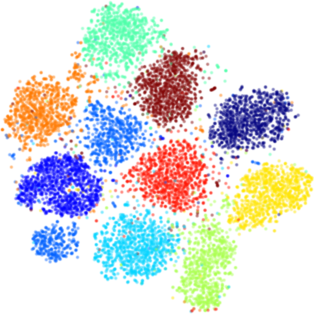}
    \caption{Normal prior}
\end{subfigure}
\begin{subfigure}[t]{0.2\textwidth}
    \centering
    \includegraphics[width=\textwidth]{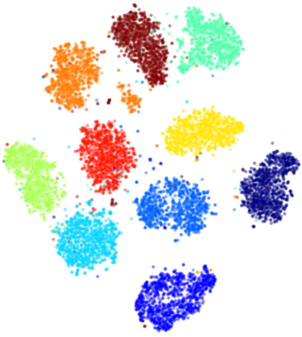}
    \caption{\acronym(200) prior}
\end{subfigure}
\caption{TSNE visualizations of the latent space of the MNIST test set with different priors,
color-coded according to class. \acronym\;prior appears to learn a space with more
separated, concentrated clusters.}
\label{fig:tsne-tmc-normal}
\end{figure}

The representations
learned with the \acronym\;
consistently achieve better accuracy,
though  in MNIST, \acronym\;prior
and MAF reach very similar test
accuracy at 100 labels per class.
The advantage of the \acronym\;prior is especially clear
in Omniglot
(\autoref{tab:semisupervised-mnist}
and \autoref{tab:semisupervised-omniglot} contain the exact numbers).
We believe our advantage in this task
comes from ability of the \acronym\;prior
to model discrete structure.
TSNE visualizations
in \autoref{fig:tsne-tmc-normal} and \autoref{fig:tsne}
indicate
clusters are more concentrated and separated
with the \acronym\;prior
than with other priors,
though TSNE visualizations should be
taken with a grain of salt.

\paragraph{Information retrieval}
We evaluated the meaningfulness of Euclidean distances
in the learned latent space
by measuring precision-recall
when querying the test set.
We take each element of the test set
and sort all other members according to
their $L_2$ distance in the latent space.
From this ranking, we produce
precision-recall curves for each 
of the query and 
plot the average precision-recall
over the entire test set in \autoref{fig:prec-rec}.
We also report the area-under-the-curve (AUC)
measure for each of these curves in \autoref{tab:auc}.

AUC numbers for Omniglot are low across the board
because of the large number of classes
and low number of instances per class.
However,
in both datasets the \acronym\;prior consistently
achieves the highest AUC,
especially with MNIST. 
The \acronym\;prior encourages
tree-distance to correspond to
squared Euclidean distance, as branch
lengths in the tree are 
variances in a Gaussian likelihoods.
We thus suspect distances in a \acronym\;prior
latent space to be more informative
and better for information retrieval.

\paragraph{Held-out log-likelihood}
We estimate held-out log-likelihoods for the four VAEs we
trained with comparable architectures and different priors.
(We exclude DVAE$\sharp$ since its architecture is substantially
different, and the classical autoencoder since it lacks generative
semantics.)
We use 1000 importance-weighted samples \citep{burda2015importance}
to estimate held-out log-likelihood,
and report the results in \autoref{tab:holl}.
We find that, although \acronym\;outperforms the other
priors on downstream tasks, it only achieves
middling likelihood numbers.
This result is consistent with the findings of \citet{chang2009reading} that held-out log-likelihood is not necessarily correlated with interpretability or usefulness for downstream tasks.
\begin{table}[H]
\centering
\input{results/ir.tex}
\caption{Averaged precision-recall AUC on MNIST/Omniglot test datasets}
\label{tab:auc}
\end{table}

\begin{table}[H]
\centering
\begin{tabular}{r|cc}
\toprule
Prior & MNIST & Omniglot\\ \midrule
Normal    & -83.789 & -89.722\\
MAF       & \textbf{-80.121} & \textbf{-86.298}\\
Vamp & -83.0135 & -87.604\\
\acronym & -83.401 & -87.105\\
\bottomrule
\end{tabular}
\caption{MNIST/Omniglot test log-likelihoods}
\label{tab:holl}
\end{table}
\section{Discussion}
Learning discrete, hierarchical structure in a latent space
opens a new opportunity:
interactive deep unsupervised learning.
User-provided constraints have been used
in both flat and
hierarchical clustering \citep{wagstaff2000clustering,awasthi2014local}, so an interesting
follow up to this work would be incorporating
constraints into the \acronym\; prior, as
in \citet{vikram2016interactive}, which could potentially
enable user-guided representation learning.

\bibliographystyle{apalike}
\bibliography{main}

\include{supplement}

\end{document}

%% file: defs.tex
\usepackage{xcolor}

\newcommand{\N}{\mathcal{N}}
\newcommand{\E}{\mathbb{E}}

\renewcommand{\L}{\mathcal{L}}

\newcommand{\papertitle}{The LORACs prior for VAEs: Letting the Trees Speak for the Data}
\newcommand{\acronym}{LORACs}
\newcommand{\acronymexplanation}{Latent ORganization of Arboreal Clusters}
\newcommand{\acronympronunciation}{lorax}

\newcommand{\iid}[1]{\stackrel{\mathrm{iid}}{#1}}
\newcommand{\given}{\,|\,}

%% file: results/ir.tex
\begin{tabular}{r|cc}
\toprule
     Prior &  MNIST & Omniglot\\
\midrule
  No prior & 0.429 & 0.078\\
    Normal & 0.317 & 0.057\\
 VAMP & 0.502 & 0.063\\
     DVAE\# & 0.490 & 0.024\\
       MAF & 0.398 & 0.070\\
 \acronym & \textbf{0.626} & \textbf{0.087}\\
\bottomrule
\end{tabular}

%% file: supplement.tex
\renewcommand{\theequation}{\thesection.\arabic{equation}}
\renewcommand{\thetable}{\thesection.\arabic{table}}
\renewcommand{\thefigure}{\thesection.\arabic{figure}}

\onecolumn

\aistatstitle{\papertitle\;- Supplement}

\appendix

\section{Additional visualizations}

\begin{figure}[h]
\begin{subfigure}[t]{0.3\textwidth}
    \centering
    \includegraphics[width=\textwidth]{img/mnist/tsne/mnist2-tsne-normal.png}
    \caption{Normal prior}
\end{subfigure}
\begin{subfigure}[t]{0.3\textwidth}
    \centering
    \includegraphics[width=\textwidth]{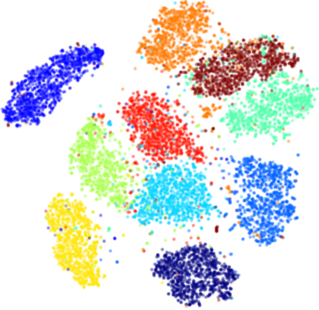}
    \caption{No prior}
\end{subfigure}
\begin{subfigure}[t]{0.3\textwidth}
    \centering
    \includegraphics[width=\textwidth]{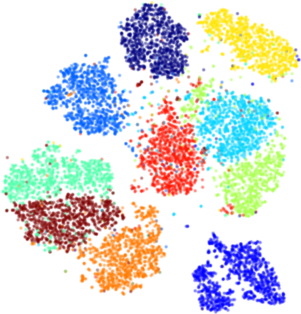}
    \caption{Vamp(500) prior}
\end{subfigure}
~
\begin{subfigure}[t]{0.3\textwidth}
    \centering
    \includegraphics[width=\textwidth]{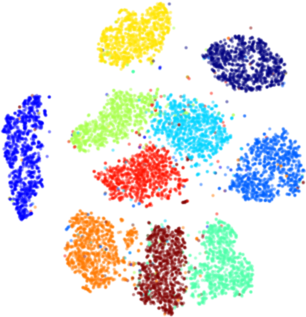}
    \caption{MAF prior}
\end{subfigure}
\begin{subfigure}[t]{0.3\textwidth}
    \centering
    \includegraphics[width=\textwidth]{img/mnist/tsne/mnist2-tsne-tmc.png}
    \caption{\acronym(200) prior}
\end{subfigure}
\caption{TSNE visualizations of the latent space of the MNIST test set with various prior distributions, color-coded according to class.}
\label{fig:tsne}
\end{figure}

\begin{figure}[h]
    \centering
    \includegraphics[width=0.5\textwidth]{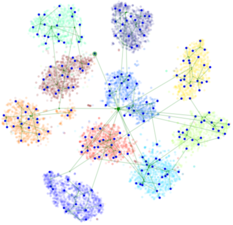}
    \caption{A TSNE visualization of the latent space for the TMC(200) model with inducing points and one sample from $q(\tau; s_{1:M})$ plotted. Internal nodes are visualized by computing their expected posterior values, and branches are plotted in 2-d space.}
    \label{fig:mnist-tsne-tree}
\end{figure}

\begin{figure}[h]
\centering
\includegraphics[width=0.5\textwidth]{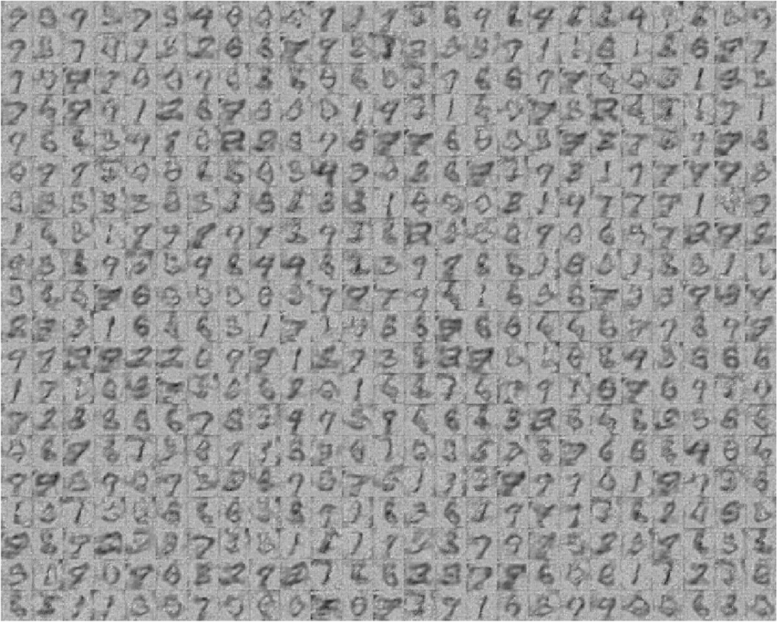}
\caption{MNIST VampPrior learned pseudo-inputs.}
\label{fig:mnist-vamp-inducing}
\end{figure}
\begin{figure}[h]
\centering
\includegraphics[width=0.5\textwidth]{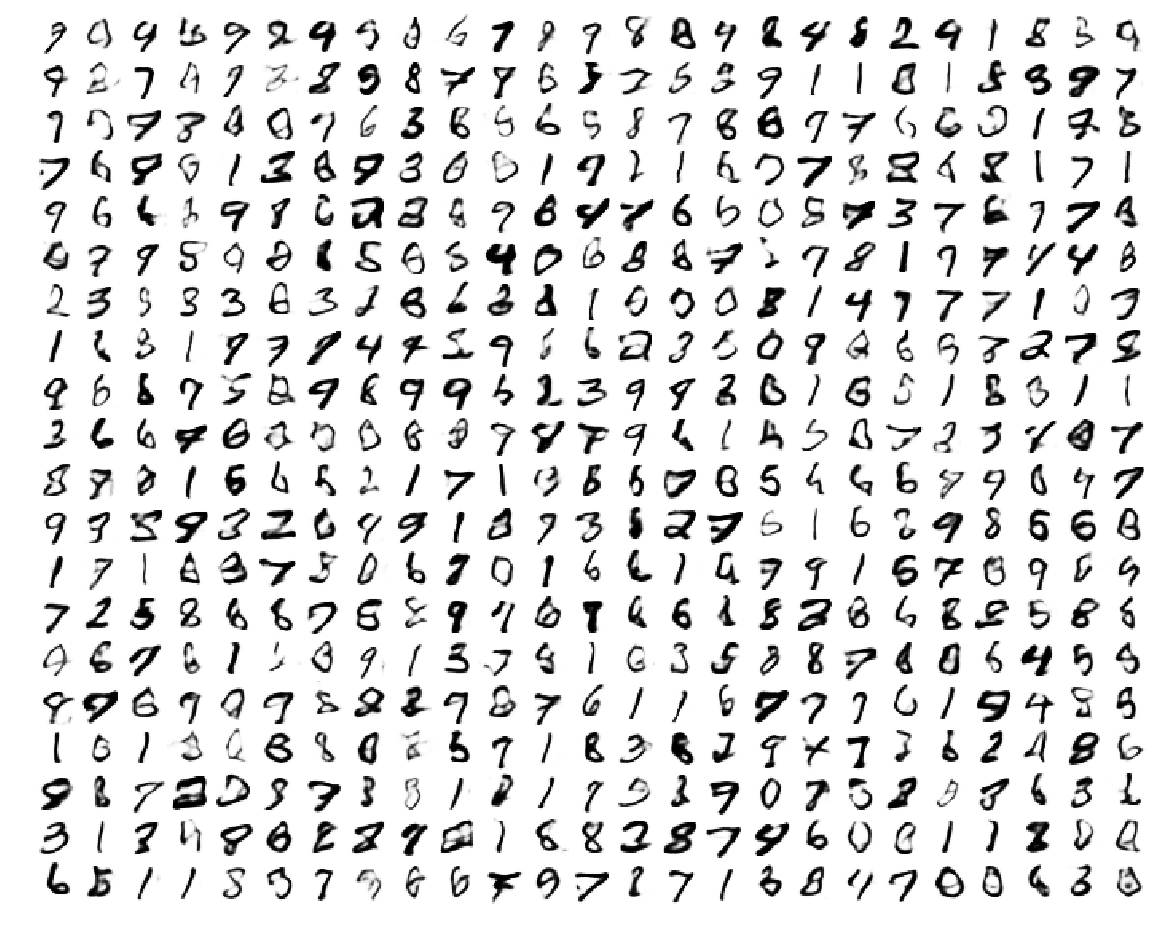}
\caption{MNIST VampPrior reconstructed pseudo-inputs obtained by
deterministically encoding and decoding each pseudo-input.}
\label{fig:mnist-vamp-inducing-outputs}
\end{figure}

\begin{figure}[h]
\centering
\includegraphics[width=0.4\textwidth]{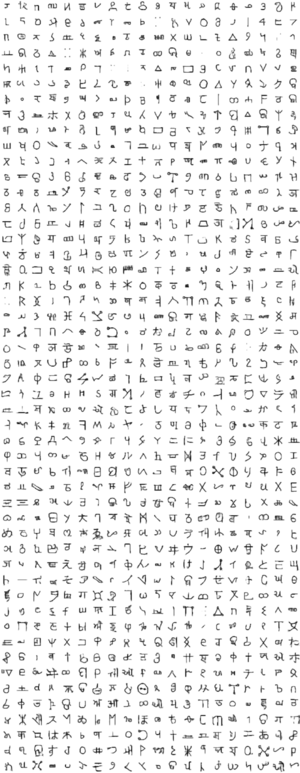}
\caption{Omniglot learned inducing points.}
\label{fig:omniglot-inducing-points}
\end{figure}

\begin{figure}[h]
\centering
\includegraphics[width=0.8\textwidth]{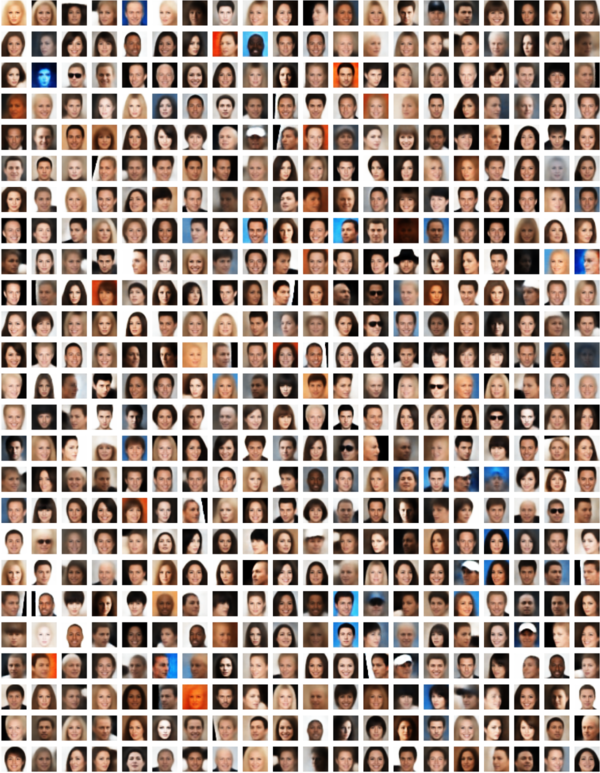}
\caption{CelebA learned inducing points.}
\label{fig:celeba-inducing-points}
\end{figure}

\begin{landscape}
\section{Empirical results}
\thispagestyle{empty}

\begin{table}[H]
\scriptsize
\centering
\setlength{\tabcolsep}{2pt}
\input{results/cf-mnist.tex}
\caption{MNIST few-shot classification results.}
\label{tab:semisupervised-mnist}
\end{table}

\begin{table}[H]
\scriptsize
\setlength{\tabcolsep}{2pt}
\centering
\input{results/cf-omniglot.tex}
\caption{Omniglot few-shot classification results.}
\label{tab:semisupervised-omniglot}
\end{table}
\begin{table}[H]
\scriptsize
\setlength{\tabcolsep}{2pt}
\begin{tabular}{r|cc}
\toprule
     \# of inducing points &  200 & 500\\
\midrule
        & 0.9428 & 0.9474\\
\bottomrule
\end{tabular}
\centering
\caption{MNIST few-shot classification with labeled inducing points.}
\label{tab:inducing-point-labels}
\end{table}
\end{landscape}

\begin{figure}[H]
\centering
\begin{subfigure}[h]{0.4\textwidth}
    \centering
    \includegraphics[width=\textwidth]{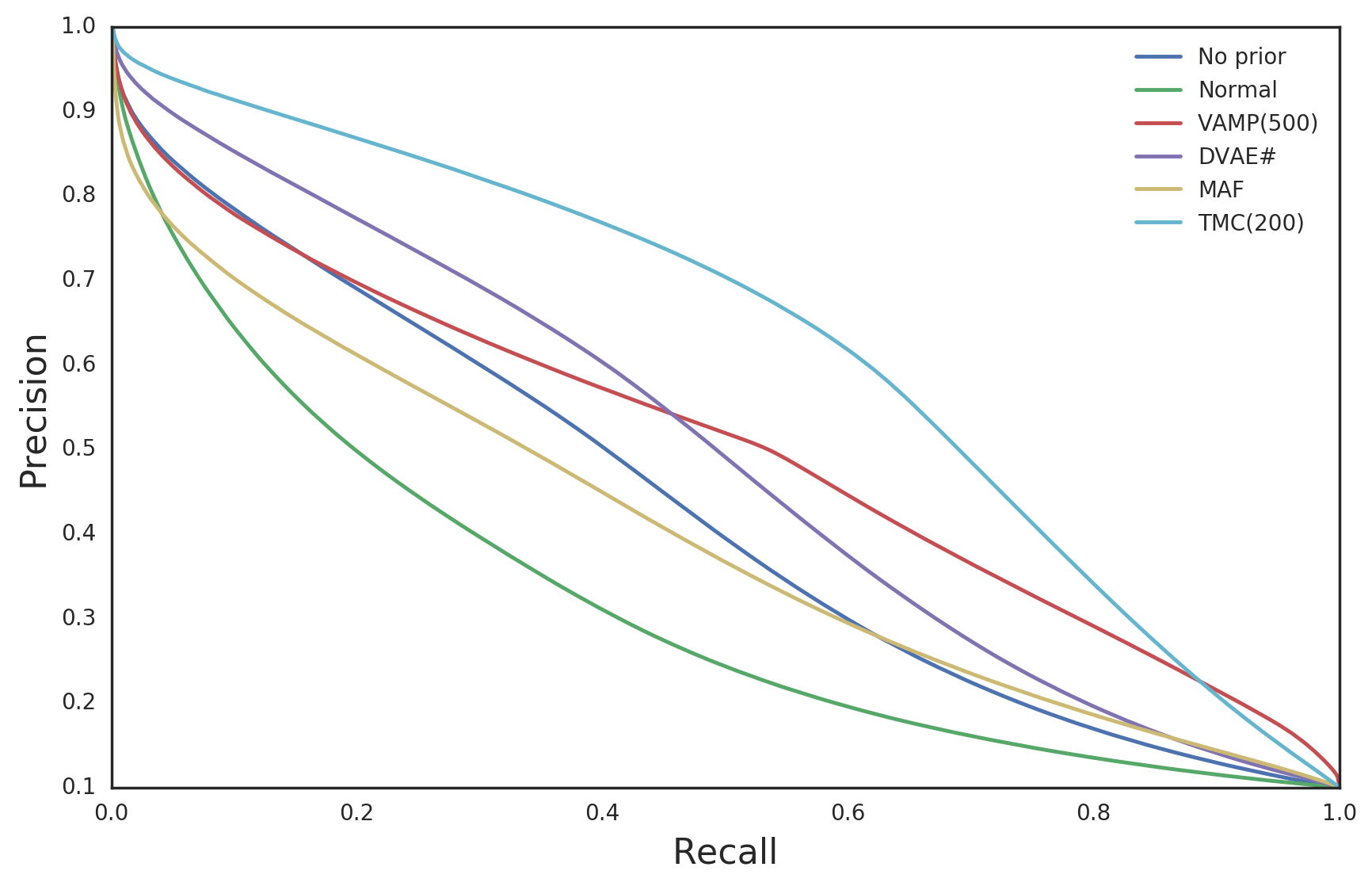}
    \caption{MNIST}
\end{subfigure}
\begin{subfigure}[h]{0.4\textwidth}
    \centering
    \includegraphics[width=\textwidth]{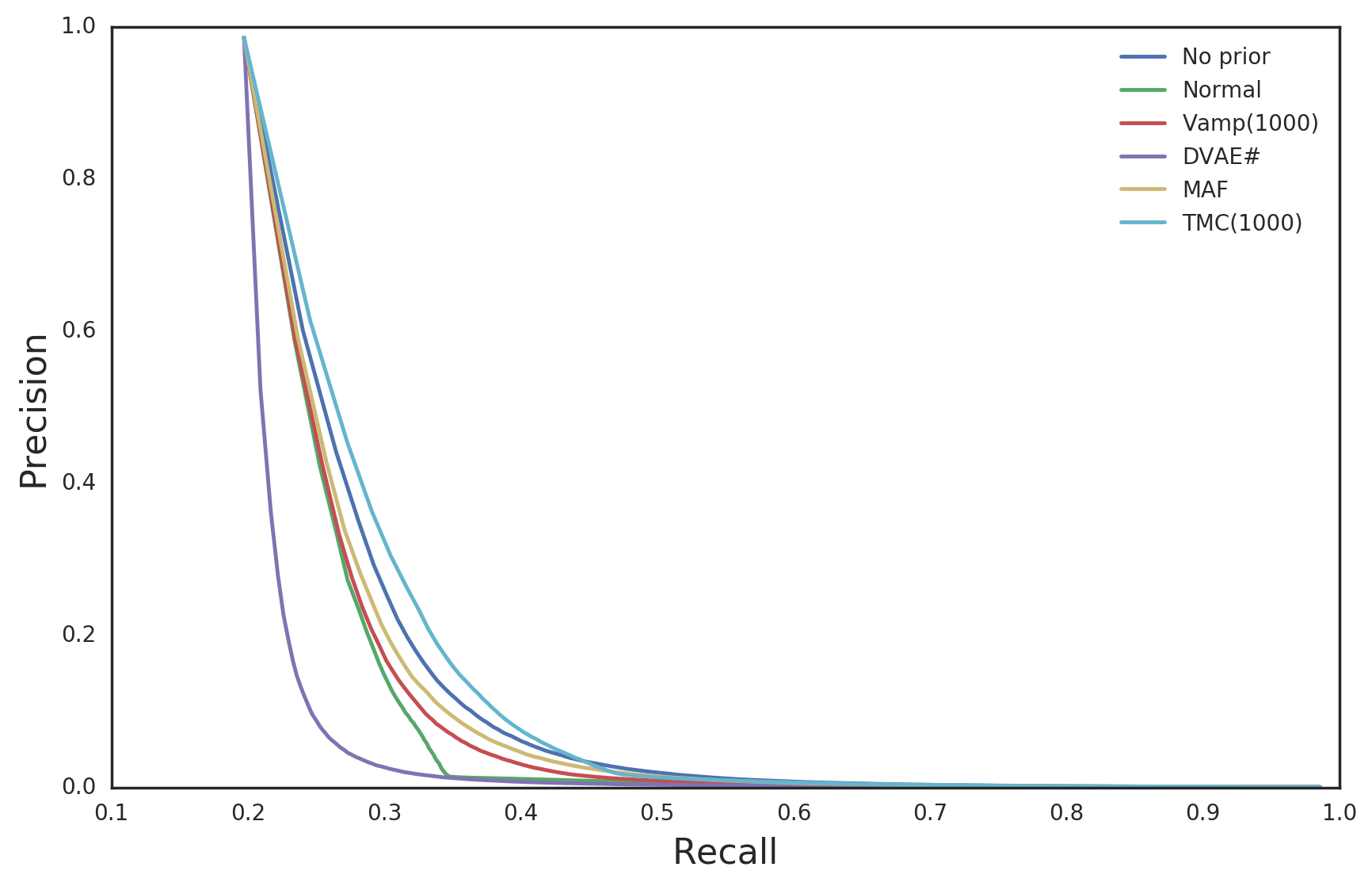}
    \caption{Omniglot}
\end{subfigure}
\caption{Averaged precision-recall curves over test datasets.}
\label{fig:prec-rec}
\end{figure}

\section{Algorithm details}
\label{sec:algorithm-details}

\subsection{Stick breaking process}
\label{sec:stick-breaking}
Consider inserting a node $N + 1$ into the tree in between vertices $u$ and $v$
such that $t_v > t_u$,
creating branch $e_{N + 1}$.
The inserted node has time $t_{N + 1}$ with probability according to the stick breaking process, i.e.
\begin{equation}
\textstyle r(t_{N+1}\given e_{N+1}, V, E) =
\mathrm{Beta}\left(\frac{t_v - t_{N + 1}}{1 - t_{N + 1}}; a, b\right)\mathrm{Beta}\left(\frac{t_{N + 1} - t_u}{1 - t_u}; a, b\right).
\end{equation}

\subsection{Belief propagation in TMCs}

The TMC is at the core of the \acronym\;
prior.
Recall that the TMC is
a prior over phylogenies $\tau$,
and after attaching a Gaussian random walk (GRW),
we obtain a distribution over $N$ vectors in $\mathbb{R}^d$,
corresponding to the leaves, $r(z_{1:N} \given \tau)$.
However, the GRW samples
latent vectors at internal nodes $z_{V_{\text{int}}}$.
Rather than explicitly representing these values,
in this work we marginalize them out, i.e.
\begin{equation}
    r(z_{1:N} \given \tau) = \int r(z_{1:N} \given z_{V_{\text{int}}}, \tau)p(z_{V_{\text{int}}} \given \tau) dz_{V_{\text{int}}}
\end{equation}
This marginalization process can be done efficiently, because our graphical model
is tree-shaped and all nodes have Gaussian likelihoods. Belief propagation is a message-passing
framework for marginalization and we utilize message-passing
for several TMC inference queries. The main queries we are interested in are:
\begin{enumerate}
    \item $r(z_{1:N}, \tau)$ - for the purposes of MCMC, we are interested in computing the joint likelihood of a set of observed
    leaf values and a phylogeny.
    \item $r(z_n \given z_{\backslash n}, \tau)$ - this query computes the posterior density over one leaf given all the others;
    we use this distribution when computing the posterior predictive density of a TMC.
    \item $\nabla_{z_{\backslash n}} r(z_n \given z_{\backslash n}, \tau)$ - this query is the gradient of the predictive density 
    of a single leaf with respect to the values at all other leaves. This query is used
    when computing gradients of the ELBO w.r.t $s_{1:M}$ in the \acronym\;prior.
\end{enumerate}

\paragraph{Message passing} Message
passing treats the tree as an undirected graph. 
We first pick start node $v_{\text{start}}$
and request messages from each of $v_{\text{start}}$'s neighbors.

Message passing is thereafter defined recursively. 
When a node $v$ has requested messages from a source node $s$,
it thereafter requests messages from all its neighbors but $s$.
The base case for this recursion is a leaf node $v_n$,
which returns a message with the following contents:
\begin{equation}
    \nu_n = \bm{0};\quad \mu_n = z_n;\quad \log Z_n = 0;\quad \nabla_{\nu_n}(\nu) = \bm{1};\quad \nabla_{\nu_n}(\mu) = \bm{0};\quad \nabla_{\mu_n}(\mu) = \bm{1}
\end{equation}
where bold numbers $\bm{0}\triangleq (0,\ldots,0)^\top$ and $\bm{1}\triangleq (1,\ldots,1)^\top$ denote vectors obtained by repeating a scalar $d$ times.

In the recursive case, consider being at a node $i$ and receiving a set of messages from its neighbors $M$.
\begin{equation}
\begin{split}
    \nu_i = \frac{1}{\sum_{m \in M}\frac{1}{\nu_m + e_{im}}} ;\quad \mu_i = v_i \sum_{m \in M} \frac{\mu_m}{\nu_m + e_{im}}\\
\end{split}
\end{equation}
where $e_{im}$ is the length of the edge between nodes $i$ and $m$.
These messages are identical to those used in \cite{boyles2012time}.

Additionally, our messages include gradients w.r.t. \emph{every} leaf node
downstream of the message.
We update each of these gradients when computing the new message
and pass them along to the source node.
Gradients with respect to one of these nodes $j$ are calculated as
\begin{equation}
\begin{split}
    \nabla_{\nu_j}(\nu) &= \nabla_{\nu_j} \nu_i\\
    \nabla_{\nu_j}(\mu) &= \nabla_{\nu_j} \mu_i\\
    \nabla_{\mu_j}(\mu) &= \nabla_{\mu_j} \mu_i\\
\end{split}
\end{equation}
The most complicated message is the $\log Z_i$ message, which depends
on the number of incoming messages. $v_{\text{start}}$ gets
three incoming messages,
all other nodes get only two. Consider
two messages from nodes $v_k$ and $v_l$:
\begin{equation}
\begin{split}
    \Sigma_i &\triangleq (\nu_k + e_{ik} + \nu_l + e_{il})I \\
    \log Z_i &= -\frac{1}{2}\|\mu_k - \mu_l\|^2_{\Sigma_i} - \frac{1}{2}\left(\log|\Sigma_i| _ d\log2\pi\right)
\end{split}
\end{equation}
For three messages from nodes
$v_k$, $v_l$, and $v_m$:
\begin{equation}
\begin{split}
    \Sigma_i &\triangleq \left((\nu_k + e_{ik})(\nu_l + e_{il}) + (\nu_l + e_{il})(\nu_m + e_{im}) + (\nu_m + e_{im})(\nu_k + e_{ik})\right)I \\
    \log Z_i &= -\frac{1}{2}\left((\nu_m + e_{im})\|\mu_k - \mu_l\|^2_{\Sigma_i} + (\nu_k + e_{ik})\|\mu_l - \mu_m\|^2_{\Sigma_i} + (\nu_l + e_{il})\|\mu_m - \mu_k\|^2_{\Sigma_i}\right) - \frac{1}{2}\log|\Sigma_i| - \log2\pi
\end{split}
\end{equation}

With these messages, we can answer all the aforementioned inference queries.
\begin{enumerate}
    \item We can begin message passing at any internal node and compute: $\log r(z_{1:N}, \tau) = \sum_{v \in V} \log Z_v$
    \item We start message passing at $v_n$. $r(z_n \given z_{\backslash n}, \tau)$ is a Gaussian with mean $\mu_n$ and variance $\nu_n$.
    \item $\nabla_{z_{\backslash n}} r(z_n \given z_{\backslash n}, \tau)$ is $\nabla_{z_{\backslash n}} \N(z_n \given \mu_n, \nu_n I)$,
    which in turn utilizes gradients sent via message passing.
\end{enumerate}

\paragraph{Implementation} We chose
to implement the TMC and message
passing in Cython because we found raw Python to be
too slow due to function call and type-checking
overhead. Furthermore, we used diagonal rather than
scalar variances in the message passing implementation
to later support diagonal variances handed 
from the variational posterior over $z_n$.

\subsection{Variational inference for the \acronym\;prior}
\label{sec:inference-details}

The \acronym\; prior involves first sampling a tree
from the posterior distribution over TMCs
with $s_{1:M}$ as leaves.
We then sample a branch and time
for each data $z_n$ according
to the posterior predictive
distribution described in \autoref{sec:bnhc}.
We then sample a $z_n$ from
the distribution induced by the GRW likelihood model.
Finally, we pass the sampled $z_n$
through the decoder.
\begin{equation}
    \begin{split}
        \tau &\sim p(\tau ; s_{1:M}) \\
        e_n, t_n &\sim p(e_n, t_n | \tau) \\
        z_n | e_n, t_n, \tau &\sim p(z_n | e_n, t_n, \tau; s_{1:M}) \triangleq r(s_{M + 1} = z_n | e_n, t_n, \tau) \\
        x_n | z_n &\sim p_\theta(x_n | z_n)
    \end{split}
\end{equation}

Consider sampling the optimal $q^*(\tau; s_{1:M})$.

\begin{equation}
\begin{split}
    q^*(\tau; s_{1:M}) 
    &\propto \exp\{\E_q\left[\log p(\tau, z_{1:N}, x_{1:N})\right]\} \\
    &\propto \exp\{\log p(\tau ; s_{1:M}) + \sum_n \E_q\left[p(z_n | e_n, t_n, \tau)\right]\} \\
    &\propto \exp\{\log \mathrm{TMC}_N(\tau; a, b) + \sum_{m = 1}^M \log r(s_m | s_{1:m - 1}, \tau) \\&+ \sum_n \E_q\left[\log p(z_n | e_n, t_n, \tau)\right]\} \\
\end{split}
\end{equation}
We  set $q(\tau; s_{1:M}) = r(\tau \given s_{1:M})$.
We use additional variational factors
$q(e_n)$,
$q_{\xi}(t_n | e_n, z_n; s_{1:M})$,
and
$q_\phi(z_n | x_n)$.
$q_{\xi}(t_n | e_n, z_n; s_{1:M})$ is a
recognition network that outputs
the attach time for a particular branch.
Since the $q(\tau; s_{1:M})$ and $p(\tau; s_{1:M})$ terms
cancel out, we obtain the following ELBO.
\begin{equation}
    \begin{split}
    \L[q] &\triangleq \E_q \left[\log \frac{\prod_n p(e_n, t_n | \tau) p(z_n | e_n, t_n, \tau; s_{1:M})p_\theta(x_n | z_n)}{\prod_n q(e_n)q_{\xi}(t_n | e_n, z_n; s_{1:M})q_\phi(z_n | x_n)}\right] \\
    \end{split}
\end{equation}

\paragraph{Inference procedure}
In general, $q(\tau; s_{1:M})$ can
be sampled using vanilla SPR Metropolis-Hastings,
so samples from this distribution are readily available. 

For each data in the minibatch $x_n$, we pass it
through the encoder to obtain $q(z_n | x_n)$.
We then compute
\begin{equation}
    q^*(e_n) = \exp\left\{\E_q\left[\log p(e_n | t_n, z_n, \tau; s_{1:M})\right]\right\}
\end{equation}
This quantity is computed by looping
over every branch $b$ of 
a sample from $q(\tau)$,
storing incoming messages at each node,
passing the $\mu$ and $\nu$
and a sample from $q(z_n | x_n)$
into $q_\xi(t_n | e_n, s_{1:M}, z_n)$,
outputting a logistic-normal distribution
over times for that branch. We sample that
logistic normal to obtain a time $t$
to go with branch $b$. We can then compute
the log-likelihood of $z_n$ if it were
to attach to $b$ and $t$, using TMC
inference query \#2.
This log-likelihood is added to the TMC prior log-probability
of the branch being selected to obtain
a joint probability $\E_q\left[\log p(e_n)p(t_n)p(z_n | e_n, t_n, \tau; s_{1:M})\right]$ over the branch.
After doing this for every branch, we normalize
the joint likelihoods to obtain
the optimal categorical distribution over every branch
for $z_n$, $q^*(e_n)$. We then sample this
distribution to obtain an attach location and time
$e_n, t_n$ for each data in the minibatch.

The next stage is to compute gradients w.r.t. to the
learnable parameters of the model ($\theta$, $s_{1:M}$, $\phi$, and $\xi$).
In the process of calculating $q^*(e_n)$,
we have obtained samples from its corresponding 
$q_\xi(t_n | e_n, z_n, \tau; s_{1:M})$ and $q(z_n | x_n)$.
We plug these into the ELBO and can compute
gradients via automatic differentiation w.r.t. $\phi$,
$\theta$, and $\xi$. Computing gradients w.r.t.
$s_{1:M}$ is more tricky. We first examine the ELBO.

\begin{equation}
    \L[q] = \E_q \left[\log \frac{\prod_n p(e_n|\tau) p(t_n) p(z_n | e_n, t_n, \tau; s_{1:M})p_\theta(x_n | z_n)}{\prod_n q(e_n)q_{\xi}(t_n | e_n, z_n; s_{1:M})q_\phi(z_n | x_n)}\right]
\end{equation}

Consider the gradient of the ELBO with respect to $s_{1:M}$.

\begin{equation}
    \begin{split}
    \nabla_{s_{1:M}}\L[q] &=
    \nabla_{s_{1:M}}\E_q \left[\log \frac{\prod_n p(e_n|\tau) p(t_n) p(z_n | e_n, t_n, \tau; s_{1:M})p_\theta(x_n | z_n)}{\prod_n q(e_n)q_{\xi}(t_n | e_n, z_n; s_{1:M})q_\phi(z_n | x_n)}\right] \\
    &= \nabla_{s_{1:M}}\sum_{\tau} q(\tau; s_{1:M})\E_q \left[\log \frac{\prod_n p(e_n|\tau) p(t_n) p(z_n | e_n, t_n, \tau; s_{1:M})p_\theta(x_n | z_n)}{\prod_n q(e_n)q_{\xi}(t_n | e_n, z_n; s_{1:M})q_\phi(z_n | x_n)}\right]\\
    &= \sum_\tau q(\tau; s_{1:M})\nabla_{s_{1:M}}\E_q\left[\log \frac{\prod_n p(e_n|\tau) p(t_n) p(z_n | e_n, t_n, \tau; s_{1:M})p_\theta(x_n | z_n)}{\prod_n q(e_n)q_{\xi}(t_n | e_n, z_n; s_{1:M})q_\phi(z_n | x_n)}\right]\\
    &+ \sum_\tau \left(\nabla_{s_{1:M}} q(\tau; s_{1:M})\right)\E_q \left[\log \frac{\prod_n p(e_n|\tau) p(t_n) p(z_n | e_n, t_n, \tau; s_{1:M})p_\theta(x_n | z_n)}{\prod_n q(e_n)q_{\xi}(t_n | e_n, z_n; s_{1:M})q_\phi(z_n | x_n)}\right]\\
    &= \sum_\tau q(\tau; s_{1:M})\nabla_{s_{1:M}}\E_q\left[\log \frac{\prod_n p(e_n|\tau) p(t_n) p(z_n | e_n, t_n, \tau; s_{1:M})p_\theta(x_n | z_n)}{\prod_n q(e_n)q_{\xi}(t_n | e_n, z_n; s_{1:M})q_\phi(z_n | x_n)}\right]\\
    &+ \sum_\tau \left(q(\tau;s_{1:M})\nabla_{s_{1:M}} \log q(\tau; s_{1:M})\right)\E_q \left[\log \frac{\prod_n p(e_n|\tau) p(t_n) p(z_n | e_n, t_n, \tau; s_{1:M})p_\theta(x_n | z_n)}{\prod_n q(e_n)q_{\xi}(t_n | e_n, z_n; s_{1:M})q_\phi(z_n | x_n)}\right]\\
    &= \E_{q(\tau)}\left[\nabla_{s_{1:M}}\E_q\left[\log \frac{\prod_n p(e_n|\tau) p(t_n) p(z_n | e_n, t_n, \tau; s_{1:M})p_\theta(x_n | z_n)}{\prod_n q(e_n)q_{\xi}(t_n | e_n, z_n; s_{1:M})q_\phi(z_n | x_n)}\right]\right.\\
    &+ \left.\nabla_{s_{1:M}} \log q(\tau; s_{1:M})\E_q \left[\log \frac{\prod_n p(e_n|\tau) p(t_n) p(z_n | e_n, t_n, \tau; s_{1:M})p_\theta(x_n | z_n)}{\prod_n q(e_n)q_{\xi}(t_n | e_n, z_n; s_{1:M})q_\phi(z_n | x_n)}\right]\right]\\
    &= \E_{q}\left[
        \nabla_{s_{1:M}}\left(-\log q(e_n) - \log q(t_n \given z_n, e_n, \tau; s_{1:M}) + \log p(z_n \given e_n, t_n, \tau; s_{1:M})\right)
    \right] \\
    &+ \E_q\left[\nabla_{s_{1:M}}\left(\log q(\tau) + \log q(e_n)\right)\log\frac{p(e_n | \tau)}{q(e_n)}\frac{p(z_n \given z_n, e_n, t_n, \tau; s_{1:M})}{q(t_n \given z_n, e_n, \tau; s_{1:M})}\right]
    \end{split}
\end{equation}

In the last step, we expand out expectation over $e_n$ and
then pass the derivative through like we did for $\tau$.
The gradients w.r.t. $q(e_n)$ are zero, since $q^*(e_n)$ is
a partial optimum of the ELBO and we are left with:.
\begin{equation}
    \begin{split}
        \nabla_{s_{1:M}}\L[q] &= \E_q\left[\nabla_{s_{1:M}}\log p(z_N | e_n, t_n, \tau; s_{1:M})\right] - \E_q\left[\log q(t_n \given z_n, e_n, \tau; s_{1:M})\right] \\
        &+ \E_q\left[\nabla_{s_{1:M}} \log q(\tau; s_{1:M})\log\frac{p(e_n | \tau)}{q(e_n)}\frac{p(z_n \given z_n, e_n, t_n, \tau; s_{1:M})}{q(t_n \given z_n, e_n, \tau; s_{1:M})}\right]
    \end{split}
\end{equation}
The first term of the gradient is the expected gradient
of the posterior predictive density w.r.t $s_{1:M}$.
This can be calculated by using TMC inference query \#3
using samples from $q(e_n)$ and $q(t_n \given z_n, e_n, \tau; s_{1:M})$. The second term also uses the same gradients, by means
of the chain rule to differentiate through the time-amortization network.
The third term of this gradient is a score function gradient,
which we decide to not use due to the high-variance nature of score function gradients. We found that we were able to obtain strong results even with biased gradients.

\section{Details of experiments}
\label{sec:implementation-details}

We implemented the \acronym\;prior in Tensorflow
and Cython. 
For MNIST and Omniglot, our architectures are in \autoref{tab:mnist-arch} and CelebA is in \autoref{tab:celeba-arch}.

\begin{table}[H]
\centering
\begin{subfigure}[h]{0.4\textwidth}
\begin{tabular}{lll}
\toprule
Layer type & Shape \\
\midrule
Conv + ReLU & [3, 3, 64], stride 2 \\
Conv + ReLU & [3, 3, 32], stride 1 \\
Conv + ReLU & [3, 3, 16], stride 2 \\
FC + ReLU & 512 \\
Gaussian & 40 \\
\bottomrule
\end{tabular}
\caption{Encoder}
\end{subfigure}
\begin{subfigure}[h]{0.4\textwidth}
\begin{tabular}{lll}
\toprule
Layer type & Shape \\
\midrule
FC + ReLU & 3136 \\
Deconv + ReLU & [3, 3, 32], stride 2 \\
Deconv + ReLU & [3, 3, 32], stride 1 \\
Deconv + ReLU & [3, 3, 1], stride 2 \\
Bernoulli & \\
\bottomrule
\end{tabular}
\caption{Decoder}
\end{subfigure}
\caption{Network architectures for MNIST and Omniglot}
\label{tab:mnist-arch}
\end{table}

\begin{table}[H]
\centering
\begin{subfigure}[h]{0.4\textwidth}
\begin{tabular}{lll}
\toprule
Layer type & Shape \\
\midrule
Conv + ReLU & [3, 3, 64], stride 2 \\
Conv + ReLU & [3, 3, 32], stride 1 \\
Conv + ReLU & [3, 3, 16], stride 2 \\
FC + ReLU & 512 \\
Gaussian & 40 \\
\bottomrule
\end{tabular}
\caption{Encoder}
\end{subfigure}
\begin{subfigure}[h]{0.4\textwidth}
\begin{tabular}{lll}
\toprule
Layer type & Shape \\
\midrule
FC + ReLU & 4096 \\
Deconv + ReLU & [3, 3, 32], stride 2 \\
Deconv + ReLU & [3, 3, 32], stride 1 \\
Deconv + ReLU & [3, 3, 3], stride 2 \\
Bernoulli & \\
\bottomrule
\end{tabular}
\caption{Decoder}
\end{subfigure}
\caption{Network architectures for CelebA}
\label{tab:celeba-arch}
\end{table}

In general, we trained the model interleaving
one gradient step with 100 sampling steps for $q(\tau; s_{1:M})$.
We also found that experimenting with values of $a$ and $b$
in the TMC prior did not impact results significantly.
We initialized the networks with weights
from a VAE trained for 100 epochs and inducing points
were initialized using k-means. All parameters
were trained using Adam \citep{kingma2015adam} with a $10^{-3}$ learning rate
for an 100 epochs with learning rate decay to $10^{-5}$ for the last 20 epochs. 
Finally, we initialized trees with all node times close to 0,
to emulate a VAE prior.

\subsection{Baseline details}
All baselines were trained with the default architecture.
They were trained for 400 epochs,
with KL warmup ($\beta$ started at $10^{-2}$, and ramped up to $\beta = 1$
linearly over 50 epochs). They were trained using Adam with a
learning rate of $10^{-3}$, with a learning rate of $10^{-5}$ for
the last 80 epochs.

DVAE\# was trained using the default implementation from
\url{https://github.com/QuadrantAI/dvae}, which is hierarchical
VAE consisting of two Bernoulli latent variables, 200-dimensional each.
Each is learned via a feed-forward neural network 4-layers deep.
The default DVAE\# implementation also uses statically binarized MNIST
where we use dynamically binarized.

%% file: results/cf-mnist.tex
\begin{tabular}{llllllllllll}
\hline
Labels per class             & 1                          & 10                         & 20                         & 30                         & 40                         & 50                         & 60                         & 70                         & 80                         & 90                         & 100                        \\ \hline
No prior                     & $0.506 \pm 0.095$          & $0.781 \pm 0.045$          & $0.820 \pm 0.023$          & $0.829 \pm 0.020$          & $0.836 \pm 0.026$          & $0.839 \pm 0.021$          & $0.844 \pm 0.017$          & $0.846 \pm 0.017$          & $0.847 \pm 0.015$          & $0.843 \pm 0.015$          & $0.848 \pm 0.014$          \\
Normal                       & $0.396 \pm 0.076$          & $0.775 \pm 0.051$          & $0.838 \pm 0.020$          & $0.861 \pm 0.016$          & $0.874 \pm 0.011$          & $0.883 \pm 0.011$          & $0.886 \pm 0.011$          & $0.892 \pm 0.010$          & $0.896 \pm 0.010$          & $0.899 \pm 0.011$          & $0.901 \pm 0.008$          \\
Vamp(500)                    & $0.539 \pm 0.094$          & $0.849 \pm 0.035$          & $0.891 \pm 0.019$          & $0.905 \pm 0.013$          & $0.911 \pm 0.016$          & $0.918 \pm 0.012$          & $0.921 \pm 0.009$          & $0.925 \pm 0.008$          & $0.929 \pm 0.007$          & $0.928 \pm 0.005$          & $0.932 \pm 0.005$          \\
DVAE\#                       & $0.453 \pm 0.101$          & $0.735 \pm 0.027$          & $0.784 \pm 0.017$          & $0.801 \pm 0.012$          & $0.813 \pm 0.013$          & $0.824 \pm 0.014$          & $0.830 \pm 0.012$          & $0.835 \pm 0.011$          & $0.841 \pm 0.007$          & $0.842 \pm 0.007$          & $0.846 \pm 0.008$          \\
MAF                          & $0.530 \pm 0.113$          & $0.869 \pm 0.029$          & $0.910 \pm 0.012$          & $0.923 \pm 0.012$          & $0.930 \pm 0.007$          & $0.933 \pm 0.010$          & $0.938 \pm 0.008$          & $0.940 \pm 0.008$          & $0.942 \pm 0.006$          & $0.944 \pm 0.006$          & $\mathbf{0.946 \pm 0.005}$ \\
\acronym(200) & $\mathbf{0.670 \pm 0.120}$ & $\mathbf{0.903 \pm 0.019}$ & $\mathbf{0.923 \pm 0.011}$ & $\mathbf{0.929 \pm 0.009}$ & $\mathbf{0.934 \pm 0.006}$ & $\mathbf{0.938 \pm 0.004}$ & $\mathbf{0.939 \pm 0.005}$ & $\mathbf{0.941 \pm 0.004}$ & $\mathbf{0.943 \pm 0.004}$ & $\mathbf{0.944 \pm 0.003}$ & $0.945 \pm 0.003$          \\ \hline
\end{tabular}

%% file: results/cf-omniglot.tex
\begin{tabular}{llllll}
\toprule
Labels per class &                 1  &                 2  &                 5  &                 10 &                 15 \\
\midrule
No prior   &  $0.140 \pm 0.012$ &  $0.179 \pm 0.008$ &  $0.225 \pm 0.006$ &  $0.252 \pm 0.009$ &  $0.290 \pm 0.001$ \\
Normal     &  $0.107 \pm 0.007$ &  $0.134 \pm 0.010$ &  $0.187 \pm 0.008$ &  $0.246 \pm 0.006$ &  $0.285 \pm 0.000$ \\
Vamp(1000) &  $0.116 \pm 0.011$ &  $0.148 \pm 0.009$ &  $0.210 \pm 0.003$ &  $0.270 \pm 0.005$ &  $0.300 \pm 0.000$ \\
DVAE\#      &  $0.042 \pm 0.004$ &  $0.060 \pm 0.006$ &  $0.091 \pm 0.003$ &  $0.121 \pm 0.001$ &  $0.141 \pm 0.000$ \\
MAF        &  $0.096 \pm 0.008$ &  $0.129 \pm 0.006$ &  $0.177 \pm 0.010$ &  $0.222 \pm 0.007$ &  $0.237 \pm 0.002$ \\
\acronym(1000)  &  $\mathbf{0.173 \pm 0.005}$ &  $\mathbf{0.236 \pm 0.005}$ &  $\mathbf{0.330 \pm 0.008}$ &  $\mathbf{0.403 \pm 0.006}$ &  $\mathbf{0.441 \pm 0.000}$ \\
\bottomrule
\end{tabular}

%% file: main.bbl
\begin{thebibliography}{}

\bibitem[Awasthi et~al., 2014]{awasthi2014local}
Awasthi, P., Balcan, M., and Voevodski, K. (2014).
\newblock Local algorithms for interactive clustering.
\newblock In {\em International Conference on Machine Learning}, pages
  550--558.

\bibitem[Blei et~al., 2003]{blei2003latent}
Blei, D.~M., Ng, A.~Y., and Jordan, M.~I. (2003).
\newblock Latent {D}irichlet allocation.
\newblock {\em Journal of Machine Learning Research}, 3(Jan):993--1022.

\bibitem[Boyles and Welling, 2012]{boyles2012time}
Boyles, L. and Welling, M. (2012).
\newblock The time-marginalized coalescent prior for hierarchical clustering.
\newblock In {\em Advances in Neural Information Processing Systems}, pages
  2969--2977.

\bibitem[Burda et~al., 2015]{burda2015importance}
Burda, Y., Grosse, R., and Salakhutdinov, R. (2015).
\newblock Importance weighted autoencoders.
\newblock {\em arXiv preprint arXiv:1509.00519}.

\bibitem[Burgess et~al., 2018]{burgess2018understanding}
Burgess, C.~P., Higgins, I., Pal, A., Matthey, L., Watters, N., Desjardins, G.,
  and Lerchner, A. (2018).
\newblock Understanding disentangling in $\beta$-vae.
\newblock {\em arXiv preprint arXiv:1804.03599}.

\bibitem[Chang et~al., 2009]{chang2009reading}
Chang, J., Gerrish, S., Wang, C., Boyd-Graber, J.~L., and Blei, D.~M. (2009).
\newblock Reading tea leaves: How humans interpret topic models.
\newblock In {\em Advances in neural information processing systems}, pages
  288--296.

\bibitem[Goyal et~al., 2017]{goyal2017nonparametric}
Goyal, P., Hu, Z., Liang, X., Wang, C., Xing, E.~P., and Mellon, C. (2017).
\newblock Nonparametric variational auto-encoders for hierarchical
  representation learning.
\newblock In {\em ICCV}, pages 5104--5112.

\bibitem[Griffiths et~al., 2004]{griffiths2004hierarchical}
Griffiths, T.~L., Jordan, M.~I., Tenenbaum, J.~B., and Blei, D.~M. (2004).
\newblock Hierarchical topic models and the nested chinese restaurant process.
\newblock In {\em Advances in neural information processing systems}, pages
  17--24.

\bibitem[Hoffman and Johnson, 2016]{hoffman2016elbo}
Hoffman, M.~D. and Johnson, M.~J. (2016).
\newblock Elbo surgery: yet another way to carve up the variational evidence
  lower bound.
\newblock In {\em Workshop in Advances in Approximate Bayesian Inference,
  NIPS}.

\bibitem[Johnson et~al., 2016]{johnson2016svae}
Johnson, M.~J., Duvenaud, D., Wiltschko, A.~B., Datta, S.~R., and Adams, R.~P.
  (2016).
\newblock Composing graphical models with neural networks for structured
  representations and fast inference.
\newblock In {\em Neural Information Processing Systems}.

\bibitem[Kingma and Ba, 2015]{kingma2015adam}
Kingma, D. and Ba, J. (2015).
\newblock {Adam}: A method for stochastic optimization.
\newblock In {\em International Conference on Learning Representations}.

\bibitem[Kingma and Welling, 2014]{kingma2013autoencoding}
Kingma, D.~P. and Welling, M. (2014).
\newblock Auto-encoding variational {B}ayes.
\newblock {\em International Conference on Learning Representations}.

\bibitem[Kingman, 1982]{kingman1982coalescent}
Kingman, J. F.~C. (1982).
\newblock The coalescent.
\newblock {\em Stochastic processes and their applications}, 13(3):235--248.

\bibitem[Lake et~al., 2015]{omniglot}
Lake, B.~M., Salakhutdinov, R., and Tenenbaum, J.~B. (2015).
\newblock {Human-level concept learning through probabilistic program
  induction}.
\newblock {\em Science}, 350(6266):1332--1338.

\bibitem[LeCun and Cortes, 2010]{mnist}
LeCun, Y. and Cortes, C. (2010).
\newblock {MNIST} handwritten digit database.

\bibitem[Lin et~al., 2018]{lin2018variational}
Lin, W., Hubacher, N., and Khan, M.~E. (2018).
\newblock Variational message passing with structured inference networks.
\newblock {\em arXiv preprint arXiv:1803.05589}.

\bibitem[Liu et~al., 2015]{celeba}
Liu, Z., Luo, P., Wang, X., and Tang, X. (2015).
\newblock Deep learning face attributes in the wild.
\newblock In {\em Proceedings of International Conference on Computer Vision
  (ICCV)}.

\bibitem[Neal, 2003]{neal2003density}
Neal, R.~M. (2003).
\newblock Density modeling and clustering using dirichlet diffusion trees.
\newblock {\em Bayesian statistics}, 7:619--629.

\bibitem[Papamakarios et~al., 2017]{papamakarios2017masked}
Papamakarios, G., Murray, I., and Pavlakou, T. (2017).
\newblock Masked autoregressive flow for density estimation.
\newblock In {\em Advances in Neural Information Processing Systems}, pages
  2338--2347.

\bibitem[Radford et~al., 2015]{radford2015unsupervised}
Radford, A., Metz, L., and Chintala, S. (2015).
\newblock Unsupervised representation learning with deep convolutional
  generative adversarial networks.
\newblock {\em arXiv preprint arXiv:1511.06434}.

\bibitem[Rezende et~al., 2014]{rezende2014stochastic}
Rezende, D.~J., Mohamed, S., and Wierstra, D. (2014).
\newblock Stochastic backpropagation and approximate inference in deep
  generative models.
\newblock In {\em Proceedings of the 31st International Conference on Machine
  Learning}, pages 1278--1286.

\bibitem[Salimans et~al., 2017]{salimans2017pixelcnn++}
Salimans, T., Karpathy, A., Chen, X., Kingma, D.~P., and Bulatov, Y. (2017).
\newblock Pixelcnn++: Improving the pixelcnn with discretized logistic mixture
  likelihood and other modifications.
\newblock In {\em International Conference on Learning Representations (ICLR)}.

\bibitem[Snelson and Ghahramani, 2006]{snelson2006sparse}
Snelson, E. and Ghahramani, Z. (2006).
\newblock Sparse gaussian processes using pseudo-inputs.
\newblock In {\em Advances in neural information processing systems}, pages
  1257--1264.

\bibitem[Tipping and Bishop, 1999]{tipping1999probabilistic}
Tipping, M.~E. and Bishop, C.~M. (1999).
\newblock Probabilistic principal component analysis.
\newblock {\em Journal of the Royal Statistical Society: Series B (Statistical
  Methodology)}, 61(3):611--622.

\bibitem[Tomczak and Welling, 2018]{tomczak2018vae}
Tomczak, J. and Welling, M. (2018).
\newblock Vae with a vampprior.
\newblock In {\em International Conference on Artificial Intelligence and
  Statistics}, pages 1214--1223.

\bibitem[Vahdat et~al., 2018]{vahdat2018dvae}
Vahdat, A., Andriyash, E., and Macready, W.~G. (2018).
\newblock Dvae\#: Discrete variational autoencoders with relaxed {B}oltzmann
  priors.
\newblock In {\em Neural Information Processing Systems (NIPS)}.

\bibitem[Vikram and Dasgupta, 2016]{vikram2016interactive}
Vikram, S. and Dasgupta, S. (2016).
\newblock Interactive bayesian hierarchical clustering.
\newblock In {\em International Conference on Machine Learning}, pages
  2081--2090.

\bibitem[Wagstaff and Cardie, 2000]{wagstaff2000clustering}
Wagstaff, K. and Cardie, C. (2000).
\newblock Clustering with instance-level constraints.
\newblock {\em AAAI/IAAI}, 1097:577--584.

\end{thebibliography}
